\documentclass{article}

 \usepackage[main, final]{neurips_2026}

\usepackage[utf8]{inputenc}
\usepackage[T1]{fontenc}
\usepackage[backref=page]{hyperref}
\usepackage{url}
\usepackage{booktabs}
\usepackage{amsfonts}
\usepackage{nicefrac}
\usepackage[expansion=false]{microtype}
\PassOptionsToPackage{dvipsnames}{xcolor}
\usepackage{xcolor}

\usepackage{graphicx}
\usepackage{subfigure}
\usepackage{amsmath}
\usepackage{amssymb}
\usepackage{mathtools}
\usepackage{amsthm}
\usepackage{commands}
\theoremstyle{remark}
\usepackage{mathrsfs}

\renewcommand*{\backref}[1]{}
\renewcommand*{\backrefalt}[4]{\ifcase #1
  \or
    (cited on page~#2)\else
    (cited on pages~#2)\fi
}
\title{Graph and Simplicial Complex Prediction Gaussian Process via Hodgelet Representations}

\author{Mathieu Alain$^{1\dagger}$ \quad So Takao$^{2}$ \quad Bastian Rieck$^3$ \quad Xiaowen Dong$^4$ \quad Emmanuel Noutahi$^5$ \\
$^1$University College London \quad $^2$PhysicsX  \quad  $^3$University of Fribourg \\ $^4$University of Oxford  \quad $^5$Valence Labs \\ 
$^\dagger$\texttt{mathieu.alain.21@ucl.ac.uk}\\
}

\begin{document}

\maketitle

\begin{abstract}
Predicting labels for graph-structured data is crucial in many scientific applications. Recently, Gaussian processes with graph-level inputs have been proposed as flexible non-parametric models for classification tasks. In this work, we extend this framework to regression tasks and to simplicial complexes, enabling the modelling of attributes defined on edges and higher-order simplices. Drawing on Hodge theory, we construct representations based on the Hodge decomposition, allowing the model to distinguish between exact, co-exact, and harmonic signal components. We introduce Hodgelet representations as rich, learnable topological descriptors of simplicial complexes and show that our framework improves predictive performance across a range of applications. Our work broadens the applicability of Gaussian processes to prediction tasks at the graph and simplicial-complex levels.
\end{abstract}

\section{Introduction and motivation}

In recent years, considerable effort has been devoted to adapting machine learning methods to non-Euclidean domains such as graphs \citep{Bronstein2021, Papamarkou2024}. This setting presents numerous challenges due to the irregular structure of graphs, their varying sizes, and the information supported on different graph elements, such as vertices and edges. Predicting graph-level labels, whether continuous or discrete, arises in applications ranging from estimating molecular properties \citep{Reiser2022} to classifying textual documents \citep{Rousseau2015}.

Simplicial complexes (SCs) generalise graphs by incorporating higher-order interactions. Whereas graphs encode pairwise interactions through edges, SCs contain simplices that can represent higher-order interactions among vertices. Of particular interest are simplicial $2$-complexes, which extend graphs by including $2$-simplices, i.e., triangular faces (filled triangles) formed by triples of vertices. Among their many applications, simplicial complexes can represent two-dimensional meshes \citep{Ballester2024} and higher-order interactions in protein--protein interaction networks \citep{maloddognin2019functional}. Moreover, any graph can be extended to a simplicial $2$-complex, for example, by filling its $3$-cliques with $2$-simplices, enabling a richer topological representation \citep{bodnar2021weisfeiler}.

Graph neural networks (GNNs) are widely used for prediction tasks on graphs. However, they can require substantial amounts of training data, may be difficult to interpret despite ongoing efforts to develop explainability methods \citep{ying2019gnnexplainer,luo2020parameterized}, and can be prone to overfitting, particularly in data-scarce settings \citep{rong2020dropedge}. Increasing GNN depth can also lead to oversmoothing \citep{li2018deeper}. Furthermore, capturing global graph properties through message passing may require many layers, especially for graphs with large diameters or structural bottlenecks.

Graph kernels provide an alternative \citep{Borgwardt2020}. Despite the prevalence of GNNs, graph-kernel methods can achieve competitive performance \citep{kriege2020survey}. Although these methods can be more data-efficient and interpretable, designing effective graph kernels remains challenging. Many existing graph kernels focus on local structural patterns, can be computationally expensive for large graphs, struggle to accommodate multidimensional vertex attributes, or rely on hand-crafted features to construct representations of graphs with varying sizes.

Another option is to construct a fixed-dimensional representation of each graph and apply a standard Euclidean kernel, such as the radial basis function (RBF) kernel, also known as the squared-exponential kernel. However, this shifts the challenge to constructing a suitable representation that captures both the graph topology and its associated attributes. Such representations can then be naturally combined with Gaussian processes (GPs) \citep{Rasmussen2005} for graph-level prediction. GPs are flexible non-parametric models that provide principled predictive uncertainty and can perform well in small-to-medium data regimes. This makes them particularly attractive when training data are limited by restricted access or the cost of data acquisition. Furthermore, the parameters used to construct the graph representations can be optimised jointly with the kernel hyperparameters by maximising the marginal likelihood.

A promising approach to capturing information from graph vertex attributes while accounting for the underlying topology is to leverage techniques from graph signal processing \citep{Ortega2018}. In particular, the graph wavelet transform \citep{Hammond2011} provides a framework for analysing graph signals with localisation in both the spatial and spectral domains. In \citet{Opolka2023}, wavelet coefficients obtained from vertex attributes are aggregated into fixed-dimensional graph representations, which are then used as inputs to a GP with a standard Euclidean kernel. The wavelet parameters are optimised jointly with the kernel hyperparameters, allowing the model to emphasise task-relevant graph frequencies. Although this approach effectively incorporates vertex attributes, it does not account for attributes defined on edges, which may encode important information such as flow magnitudes, interaction strengths, or binding affinities.

This limitation motivates the use of methods from topological signal processing \citep{Barbarossa2020}, such as the simplicial wavelet transform \citep{Roddenberry2022}, to construct representations incorporating both vertex and edge attributes. By exploiting the Hodge decomposition, simplicial wavelet transforms can be applied separately to three fundamental signal components: exact, co-exact, and harmonic. These transforms yield Hodgelet coefficients that capture gradient-like, curl-like, and harmonic contributions, respectively. When the signals are simplicial cochains, these coefficients are equivariant under simplex relabellings and changes of orientation, but not invariant. By aggregating the coefficients using permutation-invariant functions and concatenating the resulting summaries across multiple filters, we construct fixed-dimensional simplicial representations, which we call Hodgelet representations. For cochains, an aggregation that is also invariant under sign changes makes these representations invariant.

Previous studies have shown that processing Hodge components separately can improve predictive performance \citep{Battiloro2023,Yang2024,Isufi2025}. The Hodge decomposition distinguishes signal components associated with different aspects of the underlying simplicial structure. In particular, the harmonic subspace reflects global topological properties, with its dimension corresponding to the number of independent $k$-dimensional holes in the simplicial complex. Modelling these components separately allows the model to exploit their distinct characteristics.

Our work extends graph-level GP representations by incorporating edge attributes and exploiting the Hodge decomposition. Across synthetic vector-field, molecular, and power-grid benchmarks, we demonstrate that our approach achieves predictive performance competitive with or better than that of GNN baselines in small-to-medium data regimes, often with better calibration, without requiring extensive feature or architectural engineering. We focus primarily on simplicial $2$-complexes, consisting of vertices, edges, and triangular faces, which suffice for the applications considered here. Nevertheless, the framework extends to simplicial complexes of arbitrary finite dimension and can be adapted to other structures equipped with suitable boundary operators, such as cellular complexes \citep{Alain2024}. Although particularly well suited to GPs, for which kernel and wavelet parameters can be optimised jointly by maximising the marginal likelihood or a variational objective, Hodgelet representations can also be used with other machine learning models. This work is the extension of the workshop paper \citep{Alain2024b}. 

\section{Simplicial complex prediction with Gaussian processes}

In this section, we briefly introduce the concept of a simplicial complex and establish the framework for predicting its label using Gaussian processes. 

\subsection{Simplicial complexes}
\label{subsection:SC}

A $k$-\emph{simplex} is a basic building block of a \emph{simplicial complex} (SC). Simplices generalise vertices, edges, and triangles to arbitrary dimensions: a $0$-simplex is a vertex, a $1$-simplex is an edge, a $2$-simplex is a filled triangle, and so on. SCs can therefore encode higher-order relations involving more than two vertices.

\begin{figure}[htbp]
    \centering
    \subfigure[Simplicial $1$-complex (graph).]{\includegraphics[width=0.32\textwidth]{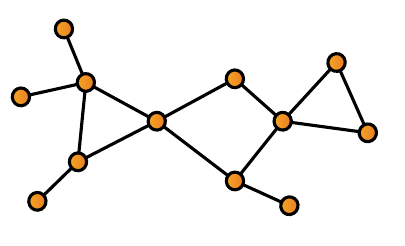}
        \label{fig:graph}
    }
    \hspace{20pt}
    \subfigure[Simplicial $2$-complex.]{\includegraphics[width=0.33\textwidth]{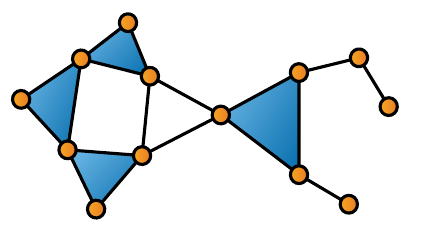}
        \label{fig:sc}
    }
    \caption{Orange vertices are $0$-simplices, black edges are $1$-simplices, and blue triangles are $2$-simplices.}
    \label{fig:simplicial-complex}
\end{figure}

Let $K \geq 0$ denote the maximum simplex dimension considered, and let $S_k$ denote a set of $N_k \geq 0$ simplices of dimension $k$, for $0\leq k \leq K$. Of particular interest are the simplex sets $S_0$, $S_1$, and $S_2$, corresponding to vertices, edges, and triangular faces, respectively. We assume that the sets $S_0, \ldots, S_K$ form a simplicial complex of dimension at most $K$ (see Figure~\ref{fig:sc}). Such sets must satisfy compatibility conditions to define a simplicial complex; we refer the reader to Appendix~\ref{appendix:sc} and \citet{Barbarossa2020} for further details. A graph (see Figure~\ref{fig:graph}) can be viewed as a simplicial $1$-complex.

Furthermore, for each simplex dimension $k$, we assume that every $k$-simplex is associated with a real-valued vector of dimension $D_k\geq 1$, specified by an attribute function defined on the simplex set $S_k$ and taking values in $\R^{D_k}$. These attributes can be interpreted as \emph{signals} on the simplices, a perspective that will be particularly useful when introducing the wavelet transform in Section~\ref{sec:RH_transform}.

Next, we fix an \emph{orientation} for each simplex. The choice of orientation is arbitrary and can, for example, be induced by a total ordering of the vertices. This allows us to encode the incidence structure of the simplicial complex, that is, how $(k-1)$-simplices are incident to $k$-simplices, using the signed \emph{incidence matrices} (also called \emph{boundary matrices}) $\bm B_k \in \Z^{N_{k-1} \times N_k}$. For notational convenience, we set $\bm B_0 \coloneqq \bm 0$ and $\bm B_{k} \coloneqq \bm 0$ for $k>K$, with these zero matrices understood to have dimensions compatible with the expressions in which they appear. After assigning an index to each simplex, the attribute functions can be represented as the matrices $\bm X_k \in \R^{N_k \times D_k}$.

We treat each attribute dimension $1 \leq d \leq D_k$ separately, as different dimensions may encode quantities of different types. For example, molecular attributes may include atomic charge, atom type, bond length, or bond type. We therefore write
$\bm X_k = [\bm x_{k1}, \ldots, \bm x_{kD_k}] \in \R^{N_k \times D_k}$,
where the vector $\bm x_{kd} \in \R^{N_k}$ contains the values of the $d$-th attribute across all $k$-simplices. For ease of exposition, we henceforth restrict attention to $D_k=1$, that is, to a scalar attribute on each $k$-simplex, and denote $\bm X_k$ simply by the attribute vector $\bm x_k \in \R^{N_k}$. We refer to $\R^{N_k}$ as the \emph{attribute space}, and represent an \emph{attributed simplicial complex} by the collection $\cal K \coloneqq \{(S_k,\bm x_k)\}_{k\leq K}$ of simplex sets and their associated attribute vectors.

Attributes defined on oriented simplices can be viewed as \emph{simplicial cochains} when their values change sign under orientation reversal. For example, reversing the orientation of an edge changes the sign of the line integral of a vector field along that edge, whereas orientation-independent attributes, such as molecular bond lengths or bond types, remain unchanged. This distinction is important for the equivariance properties of the Hodgelet coefficients introduced in Section~\ref{sec:RH_transform}.

\subsection{Gaussian process prediction of attributed SCs}

Let $\cal S$ denote a set of attributed simplicial $K$-complexes and $\cal Y$ a set of labels. Given a dataset $\cal D \coloneqq \{(\cal K_i,y_i)\}_{i\leq m} \subseteq \cal S \times \cal Y$ consisting of $m$ attributed simplicial complexes and their associated labels, our goal is to learn a predictive function $g:\cal S\to\cal Y$. We use Gaussian processes (GPs) to model this prediction problem \citep{Rasmussen2005}. GPs provide a principled Bayesian framework for learning functions by placing a probability distribution over possible functions.

We introduce a latent function $f:\cal S\to\R$ endowed with a GP prior,
\begin{equation}
f \sim \cal{GP}(\mu,\kappa),
\end{equation}
where $\mu:\cal S\to\R$ is the mean function and $\kappa:\cal S\times\cal S\to\R$ is the covariance kernel. We denote the prior distribution by $p(f)$. Both the mean function and the covariance kernel may depend on tunable hyperparameters, which often have interpretable roles, such as controlling the smoothness and variability of the latent function, in contrast to the typically less interpretable parameters of neural network models.

We further specify a likelihood $p(y_i\,|\,f(\cal K_i))$ for each $(\cal K_i,y_i)\in\cal D$. By Bayes' rule, the posterior distribution over $f$ is
\begin{equation}
p(f\,|\,\cal D)
\propto
p(f)\prod_{i=1}^m p(y_i\,|\,f(\cal K_i)).
\end{equation}
For a test simplicial complex $\cal K_*\in\cal S$, let $f_*\coloneqq f(\cal K_*)$. The posterior predictive distribution is obtained by marginalising over the posterior distribution of $f_*$:
\begin{equation}\label{eq:gp-posterior}
p(y_*\,|\,\cal K_*,\cal D)
=
\int
p(y_*\,|\,f_*)\,
p(f_*\,|\,\cal K_*,\cal D)\,df_*.
\end{equation}

For regression tasks with Gaussian observation noise, both the posterior and posterior predictive distributions admit closed-form expressions. For classification tasks, the likelihood is non-Gaussian, and the posterior is generally intractable, requiring approximate inference (see Appendix~\ref{appendix:variational_gp}). While we present the scalar-valued formulation for simplicity, multiclass classification can be accommodated by introducing multiple latent functions. The hyperparameters of the GP, together with the parameters used to construct the simplicial complex representations, can be learned jointly by maximising the marginal likelihood or an appropriate variational objective.

This describes the general GP framework; it remains to specify how the topology and attributes of a simplicial complex are incorporated into the prediction model. In the following sections, we construct a representation of each simplicial complex by applying spectral wavelet transforms to its attributes. These representations then serve as inputs to the GP.

\section{Hodge Laplacian and its decomposition} \label{sec:Hodge_Laplacian}

In this section, we introduce the discrete Hodge Laplacian and its associated decomposition. Recall from Section~\ref{subsection:SC} that, once oriented, a simplicial complex is equipped with incidence matrices $\bm B_k$, representing discrete boundary operators. Their transposes $\bm B_k^\top$ represent the corresponding discrete \emph{coboundary operators}, which are adjoint to the boundary operators under the standard Euclidean inner product. When acting on simplicial cochains, these operators can be interpreted as discrete analogues of differential operators. In particular, $\bm B_1^\top$, $\bm B_2^\top$, and $-\bm B_1$ correspond to the gradient, curl, and divergence operators, respectively \citep{Lim2020}.

\subsection{Discrete Hodge Laplacian}

In Euclidean space, the Laplace operator acting on a scalar function can be expressed as the divergence of its gradient; informally, it measures how the value of the function at a point differs from its values in the surrounding neighbourhood. On graphs, an analogous operator is given by the \emph{graph Laplacian}
\begin{equation}
   \bm L_0 \coloneqq \bm B_1 \bm B_1^\top \in \Z^{N_0 \times N_0},
\end{equation}
where $\bm B_1$ is commonly called the graph incidence matrix. This construction extends to simplicial complexes through the \emph{Hodge Laplacian} \citep{Lim2020},
\begin{equation}
    \bm L_k \coloneqq \bm L^{\mathrm{low}}_k + \bm L^{\mathrm{up}}_k \in \Z^{N_k \times N_k},
\end{equation}
where $\bm L^{\mathrm{low}}_k \coloneqq \bm B_k^\top \bm B_k$ is the \emph{lower Laplacian} and $\bm L^{\mathrm{up}}_k \coloneqq \bm B_{k+1} \bm B_{k+1}^\top$ is the \emph{upper Laplacian}. The lower Laplacian encodes interactions between $k$-simplices through $(k-1)$-simplices, whereas the upper Laplacian captures interactions between $k$-simplices through $(k+1)$-simplices.

The Hodge Laplacian is not the only Laplace-type operator; another example is the connection Laplacian \citep{Petersen2016}. However, the continuous Hodge Laplacian and its discrete counterpart have the particular advantage of inducing the Hodge decomposition, which we discuss next.

\subsection{Hodge decomposition}

\begin{figure}[htbp]
\vspace{-10pt}
    \centering
    \subfigure[Divergence-free. An edge signal is divergence-free if, at each vertex, the signed sum of the incident edge values is zero.]{
        \includegraphics[width=0.29\textwidth]{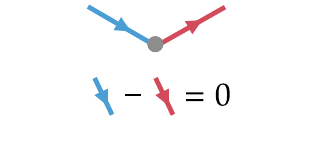}
        \label{fig:figure1}
    }
    \hspace{40pt}
    \subfigure[Curl-free. An edge signal is curl-free if, for each triangle, the signed sum of its edge values is zero.]{
        \includegraphics[width=0.31\textwidth]{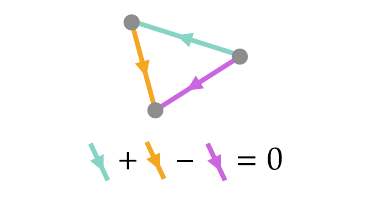}
        \label{fig:figure2}
    }
    \caption{Edge signals in the exact subspace are curl-free, whereas those in the co-exact subspace are divergence-free. Signals in the harmonic subspace are both curl-free and divergence-free.}
    \label{fig:two_figures}
\end{figure}

The \emph{Helmholtz decomposition}, often referred to as the fundamental theorem of vector calculus, states that, under suitable regularity and boundary conditions, a vector field can be decomposed into curl-free, divergence-free, and harmonic components. The curl-free component contains no rotational part (no vortices), the divergence-free component contains no sources or sinks, and the harmonic component is both curl-free and divergence-free. The discrete \emph{Hodge decomposition} provides an analogous decomposition for signals supported on simplicial complexes \citep{Lim2020}, decomposing the attribute space into a direct sum of three subspaces that are orthogonal with respect to the standard Euclidean inner product,
\begin{equation}
\R^{N_k}
=
\Im\bigl(\bm B_k^{\top}\bigr)
\oplus
\Im\bigl(\bm B_{k+1}\bigr)
\oplus
\ker\bigl(\bm L_k\bigr).
\label{HD}
\end{equation}
The subspaces $\Im(\bm B_k^{\top})$, $\Im(\bm B_{k+1})$, and $\ker(\bm L_k)$ are referred to as the \emph{exact}, \emph{co-exact}, and \emph{harmonic} subspaces, respectively (see Figure~\ref{fig:two_figures}). Collectively, these are called the \emph{Hodge subspaces}.

The Hodge decomposition also induces a partition of the eigendecomposition $\bm L_k = \bm U_k \bm \Lambda_k \bm U_k^\top$ according to the Hodge subspaces:
\begin{align}
\bm U_k
&=
\big[\bm U_{ke} \,\, \bm U_{kc} \,\, \bm U_{kh}\big]
\in \R^{N_k \times (N_{ke}+N_{kc}+N_{kh})},
\label{HDV_eigenvectors}\\
\bm \Lambda_k
&=
\mathrm{diag}\big(\bm \Lambda_{ke}, \bm \Lambda_{kc}, \bm \Lambda_{kh}\big)
\in \R^{N_k \times (N_{ke}+N_{kc}+N_{kh})},
\end{align}
where $N_{ke}$, $N_{kc}$, and $N_{kh}$ denote the dimensions of the exact, co-exact, and harmonic subspaces, respectively, with $N_k = N_{ke}+N_{kc}+N_{kh}$. Furthermore, $(\bm \Lambda_{ke}, \bm U_{ke})$ and $(\bm \Lambda_{kc}, \bm U_{kc})$ are the non-zero eigenpairs associated with the lower and upper Laplacians $\bm L_k^{\mathrm{low}}$ and $\bm L_k^{\mathrm{up}}$, respectively, while $(\bm \Lambda_{kh}, \bm U_{kh})$ are the zero eigenpairs of the Hodge Laplacian $\bm L_k$.

The value of this decomposition is that the three components carry different information and can be treated separately. In particular, the dimension of the harmonic subspace equals the number of $k$-dimensional topological holes in the simplicial complex. Moreover, a signal may be dominated by one Hodge component; for example, a vortex-dominated flow lies mostly in the co-exact subspace. Allowing a model to weight these components separately can therefore improve its predictive flexibility.

When $k=0$, the lower Laplacian $\bm L_0^{\mathrm{low}}$ is zero and the exact subspace vanishes, reflecting the absence of simplices of dimension below the vertices. Similarly, when $k=K$, the upper Laplacian $\bm L_K^{\mathrm{up}}$ is zero and the co-exact subspace vanishes because there are no $(K+1)$-simplices. For a graph, viewed as a simplicial $1$-complex, this means that vertex attributes have no exact component and edge attributes have no co-exact component. A co-exact component for edge signals can be introduced by augmenting the graph with $2$-simplices, for example, by filling its $3$-cliques to obtain a simplicial $2$-complex.

\section{Hodgelet representations for SC prediction with GPs}

We are now ready to introduce a representation of attributed SC inputs that exploits both their attributes and topology. In Section~\ref{sec:Hodge_Laplacian}, we saw that the Hodge Laplacian is a discrete analogue of a differential operator acting on the attribute space that captures the structure of an SC, including interactions between simplices of different dimensions. As we shall see, the Hodge Laplacian is also naturally related to the notion of the Fourier transform.

\subsection{Hodgelet transforms} \label{sec:RH_transform}

The \emph{Fourier transform} of a signal supported on the $k$-simplices is obtained by projecting the signal onto the eigenbasis of the Hodge Laplacian $\bm L_k$. This provides a representation of the signal in the spectral domain, making it possible to identify and manipulate specific frequencies associated with the eigenvalues of the Hodge Laplacian. Such a representation can reveal patterns that may not be apparent when examining the original signal. However, the Fourier transform is not localised in space, which limits its ability to capture spatially localised patterns.

To address this limitation, wavelet transforms provide simultaneous localisation in the spatial and spectral domains. Classical wavelet constructions can be adapted to signals supported on graphs \citep{Hammond2011} and simplicial complexes \citep{Roddenberry2022}. In our setting, this is achieved by first defining the \emph{wavelet filter}
\begin{equation}
    w_k(\lambda)
    \coloneqq
    \eta_k\,a_k\big(\alpha_k\lambda\big)
    +
    \sum_{l=1}^{J_k}
    \eta_{kl}\,b_k\big(\beta_{kl}\lambda\big),
    \qquad
    P_k
    \coloneqq
    \big\{\eta_k,\eta_{k1},\ldots,\eta_{kJ_k},\alpha_k,\beta_{k1},\ldots,\beta_{kJ_k}\big\},
\end{equation}
where $a_k:\R\rightarrow\R$ is a scaling function, $b_k:\R\rightarrow\R$ is a wavelet function, $\eta_k$ and $\eta_{kl}$ are learnable filter amplitudes, $J_k$ is the number of band-pass scales, and $P_k$ is the corresponding collection of wavelet parameters. The wavelet function acts as a band-pass filter that captures information at intermediate to high frequencies, whereas the scaling function acts as a low-pass filter that captures low-frequency information.

Rather than applying a single wavelet filter to the full spectrum $\bm\Lambda_k$, we apply separate filters to the exact, co-exact, and harmonic components $\bm\Lambda_{ke}$, $\bm\Lambda_{kc}$, and $\bm\Lambda_{kh}$, respectively. We denote these filters by $w_{ke}$, $w_{kc}$, and $w_{kh}$, with corresponding parameter sets $P_{ke}$, $P_{kc}$, and $P_{kh}$. The $k$-simplex wavelet transforms \citep{Roddenberry2022} associated with the three Hodge components, called the \emph{Hodgelet transforms}, are then defined as the matrices
\begin{align}
    \mathbf{W}_{\!P_{ke}}
    &\coloneqq
    \bm U_{ke}
    w_{ke}\big(\bm\Lambda_{ke}\big)
    \bm U_{ke}^{\top}
    \in \R^{N_k\times N_k},
    \\
    \mathbf{W}_{\!P_{kc}}
    &\coloneqq
    \bm U_{kc}
    w_{kc}\big(\bm\Lambda_{kc}\big)
    \bm U_{kc}^{\top}
    \in \R^{N_k\times N_k},
    \\
    \mathbf{W}_{\!P_{kh}}
    &\coloneqq
    \bm U_{kh}
    w_{kh}\big(\bm\Lambda_{kh}\big)
    \bm U_{kh}^{\top}
    \in \R^{N_k\times N_k}.
\end{align}
The wavelet filters $w_{k\bullet}$ are applied element-wise to the eigenvalues contained in the diagonal matrices $\bm\Lambda_{k\bullet}$.
Note that since $\bm\Lambda_{kh} = \bm 0$, the harmonic filter reduces to the constant $w_{kh}(0)$, which depends on the filter amplitudes but not on the scale parameters.

When applied to the signal $\bm x_k$, these wavelet transforms yield what we call the \emph{Hodgelet coefficients}
\begin{equation}\label{eq:hodgelet-coeffs}
\hat{\bm x}_{ke}
\coloneqq
\mathbf{W}_{\!P_{ke}}\bm x_k
\in \R^{N_k},
\qquad
\hat{\bm x}_{kc}
\coloneqq
\mathbf{W}_{\!P_{kc}}\bm x_k
\in \R^{N_k},
\qquad
\hat{\bm x}_{kh}
\coloneqq
\mathbf{W}_{\!P_{kh}}\bm x_k
\in \R^{N_k}.
\end{equation}

These coefficients are natural candidates for representing attributed simplicial complexes, as they capture information from both the attributes and the underlying structure of the complex. When the attributes are simplicial cochains, the Hodgelet coefficients are equivariant under simplex relabellings and changes of orientation, meaning that they transform consistently with these operations rather than remaining unchanged. For attributes that are not simplicial cochains, the coefficients remain equivariant under simplex relabellings but not, in general, under changes of orientation. We aggregate the coefficients to obtain fixed-dimensional representations, whose invariance properties depend on the choice of aggregation function.

\subsection{Hodgelet representations}\label{sec:representations}

We pool the Hodgelet coefficients in Equation~\eqref{eq:hodgelet-coeffs} using an \emph{aggregation function}, similarly to the aggregation operations used in message-passing neural networks \citep{gilmer2017neural}. Ideally, this function should be invariant under permutations and sign changes of the coefficients, so that, for cochains, the resulting representation is invariant under simplex relabellings and changes of orientation. We also require the aggregation function to output a scalar value. As a simple example, we consider the \emph{energy}, given by the squared $2$-norm,
\begin{equation}
    A^{}_{ke} \coloneqq \big\|\hat{\bm x}^{}_{ke}\big\|^2,
    \quad
    A^{}_{kc} \coloneqq \big\|\hat{\bm x}^{}_{kc}\big\|^2,
    \quad
    A^{}_{kh} \coloneqq \big\|\hat{\bm x}^{}_{kh}\big\|^2.
\end{equation}
The squared $2$-norm is invariant under permutations and sign changes of the coefficients, thus satisfying the desired invariance properties. Other aggregation functions may also be considered, such as the sum, minimum, and maximum, or weighted combinations of these and the energy (the \emph{mixture}), although they are not generally invariant under sign changes of the coefficients.
For example, in settings where the net value of the signal is important for prediction, the sum may provide a more appropriate aggregation than the energy.

\textbf{Non-cochain attributes.}
If the attributes do not change sign under orientation reversal (e.g., bond types), a change of orientation, given by a diagonal sign matrix $\bm S\in\R^{N_k\times N_k}$, maps the coefficients $\mathbf W_{\!P_{k\bullet}}\bm x_k$ to $\bm S\mathbf W_{\!P_{k\bullet}}\bm S\bm x_k$, and the aggregations above then generally depend on the arbitrary orientation (e.g., on the atom numbering of a molecule). For such attributes, we replace the sum and the energy by their averages over all orientations. These two \emph{orientation-averaged aggregations}, the orientation-averaged sum and energy, have the closed forms
\begin{equation}\label{eq:oavg}
\begin{aligned}
    A^{\mathrm{sum}}_{k\bullet}
    &\coloneqq
    \bb E_{\bm S}\big[\bm 1^\top\bm S\mathbf W_{\!P_{k\bullet}}\bm S\bm x_k\big]=\textstyle\sum_{i=1}^{N_k}(\mathbf W_{\!P_{k\bullet}})_{ii}\,(\bm x_k)_i,
    \\
    A^{\mathrm{energy}}_{k\bullet}
    &\coloneqq
    \bb E_{\bm S}\big[\|\bm S\mathbf W_{\!P_{k\bullet}}\bm S\bm x_k\|^2\big]=\textstyle\sum_{i=1}^{N_k}\big(\mathbf W_{\!P_{k\bullet}}^2\big)_{ii}\,(\bm x_k)_i^2,
\end{aligned}
\end{equation}
which are exactly invariant under simplex relabellings and changes of orientation (see Appendix~\ref{appendix:sc}). For such attributes, each wavelet filter contributes both values to the Hodgelet representation below, which then has twice as many entries.

Next, for each $0\leq k\leq K$, we form the \emph{Hodgelet representations} by concatenating the aggregated coefficients obtained from $F_k$ wavelet filters:
\begin{equation}
    \bm r_{k\bullet}
    \coloneqq
    \left[
        A^{(1)}_{k\bullet}
        \,\cdots\,
        A^{(F_k)}_{k\bullet}
    \right]^\top
    \in \R^{F_k},
    \qquad \bullet\in\{e,c,h\}.
\end{equation}
Using multiple wavelet filters rather than a single one enables the representation to capture information from the simplicial complex signal across multiple spectral resolutions.

\textbf{Remark 1.}
The case of $D_k>1$ attributes per $k$-simplex is handled by applying the same wavelet filters to each attribute dimension and stacking the resulting aggregations into Hodgelet representations of dimension $F_kD_k$ rather than $F_k$.

\textbf{Remark 2.}
The most expensive step in computing the Hodgelet coefficients is the eigendecomposition of $\bm L_k$, which scales cubically with the number $N_k$ of $k$-simplices. For large SCs, this cost can be reduced with Lanczos solvers or with polynomial approximations of the filters in $\bm L_k^{\mathrm{low}}$ and $\bm L_k^{\mathrm{up}}$ \citep{Defferrard2016}; the SCs in many prediction tasks are, however, small enough for a full eigendecomposition.

\subsection{Hodgelet kernel function}

Recall that $\cal S$ denotes a set of attributed simplicial complexes of dimension at most $K$. Given the Hodgelet representations, we can now define the covariance kernel $\kappa:\cal S\times\cal S\rightarrow\R$ as an additive kernel, with one term for each simplex dimension and Hodge component:
\begin{align}
    \kappa(\cal K,\cal K')
    &\coloneqq
    \sum_{k=0}^{K}
    \big(
        \sigma_{ke}^2\,\kappa_{ke}(\bm r_{ke},\bm r_{ke}')
        +
        \sigma_{kc}^2\,\kappa_{kc}(\bm r_{kc},\bm r_{kc}')
        +
        \sigma_{kh}^2\,\kappa_{kh}(\bm r_{kh},\bm r_{kh}')
    \big),
    \label{eq:add-kernel}
\end{align}
 for any simplicial complexes $\cal K,\cal K'\in\cal S$, where $\kappa_{k\bullet}:\R^{F_k}\times\R^{F_k}\rightarrow\R$, for $\bullet\in\{e,c,h\}$, are standard kernels with unit amplitude, such as the RBF kernel or kernels from the Matérn family, and the output scales $\sigma_{ke}^2,\sigma_{kc}^2,\sigma_{kh}^2>0$ weight the Hodge components; a scale close to zero effectively switches its component off. The sum over the Hodge components mirrors the Hodge decomposition \eqref{HD}, which writes a signal as the sum of its exact, co-exact, and harmonic parts. Other combinations of the component kernels are possible, such as products across simplex dimensions. The kernel hyperparameters (the output scales $\sigma_{k\bullet}^2$ and the parameters of the kernels $\kappa_{k\bullet}$) are optimised jointly with the wavelet parameters $P_{k\bullet}$ of the $F_k$ filters, which trains the model end-to-end (the full pipeline is summarised in Appendix~\ref{appendix:training_model_selection_eval}). As with standard GPs, exact training and inference have cubic time complexity $\cal O(m^3)$ in the number $m$ of training SCs, although sparse variational GPs with $M$ inducing points, which we use in all experiments, reduce it to $\cal O(m_bM^2+M^3)$ per mini-batch of size $m_b$ \citep{Titsias2009,hensman2013gaussian}. For a fixed dataset, the eigendecompositions required to construct the Hodgelet representations constitute a one-off preprocessing cost and can be computed offline.

Using separate kernels for each Hodge component provides greater modelling flexibility than using a single kernel and can improve predictive performance for edge flows \citep{Yang2024}. In addition, our construction supports continuous, multidimensional simplex attributes and SCs of varying sizes, in contrast to many conventional graph-kernel methods.

\section{Experiments}\label{sec:exps}

We consider three series of experiments (vector fields, molecules, and power grids), covering binary and multiclass classification and regression. Our aim is to demonstrate that (1) adding edge information can improve the performance of our models, (2) the Hodge decomposition offers a natural way of increasing the flexibility of our SC representations, and (3) when edge attributes are present, it is preferable to treat them as edge attributes rather than as vertex attributes on the corresponding line graph.

As baselines, we consider the graph neural network (GNN) architectures GCN \citep{Kipf2017}, GAT \citep{Velickovic2018}, GIN \citep{xu2019powerful,Hu2020}, and Graph Transformer \citep{Shi2021}; GCN uses only vertex attributes, while GAT, GIN, and Transformer use both vertex and edge attributes. Appendix~\ref{appendix:additional} adds simplicial (SNN, SCNN) and Weisfeiler--Lehman kernel (WL-SVM, WL-GP) baselines on Div-curl-free, Vortices, and MUTAG. We denote by WTGP and HTGP the GP models whose SC representations are based on the wavelet transform and on the Hodgelet transforms, respectively; the latter applies wavelet transforms to each Hodge component separately. The labels ``vertex'', ``edge'', and ``hybrid'' indicate whether a model uses the vertex attributes, the edge attributes, or both; for example, WTGP (vertex) is the GP model of \citet{Opolka2023}.
We evaluate each model over five independent repetitions (independently generated datasets, random subsamples, or random data splits). Within each repetition, we run five-fold cross-validation and compute all metrics from the pooled out-of-fold predictions; we report the mean and standard error over the repetitions. The aggregation function is fixed per benchmark (see Appendix~\ref{appendix: experiments}); the negative log predictive density (NLPD) and expected calibration error (ECE) are reported in Appendix~\ref{appendix:additional}.

\subsection{Vector field classification}
\label{exp:vectorfield}
We first conduct experiments on two synthetic datasets, which we refer to as Div-curl-free and Vortices. Both contain $100$ complexes, obtained by discretising random vector fields into edge signals on triangular meshes via the de Rham map \citep{Desbrun2008} (see Appendix~\ref{appendix:vectorfield}).
The tasks are to determine whether a vector field is predominantly divergence-free or curl-free (Div-curl-free), or whether the net circulation of a vortex-dominated fluid flow is positive or negative, i.e., anticlockwise or clockwise (Vortices). Each task is performed at different mesh resolutions, given by the number of vertices, ranging from $10$ to $150$. Within a generated Vortices dataset, all complexes share one mesh, and hence one edge orientation (see Appendix~\ref{appendix:vectorfield}). We also construct line-graph counterparts on which WTGP (vertex) and HTGP (vertex) are trained.

\begin{figure}[ht]
    \centering
    \begin{minipage}{0.49\linewidth}
        \centering
    \includegraphics[width=\linewidth]{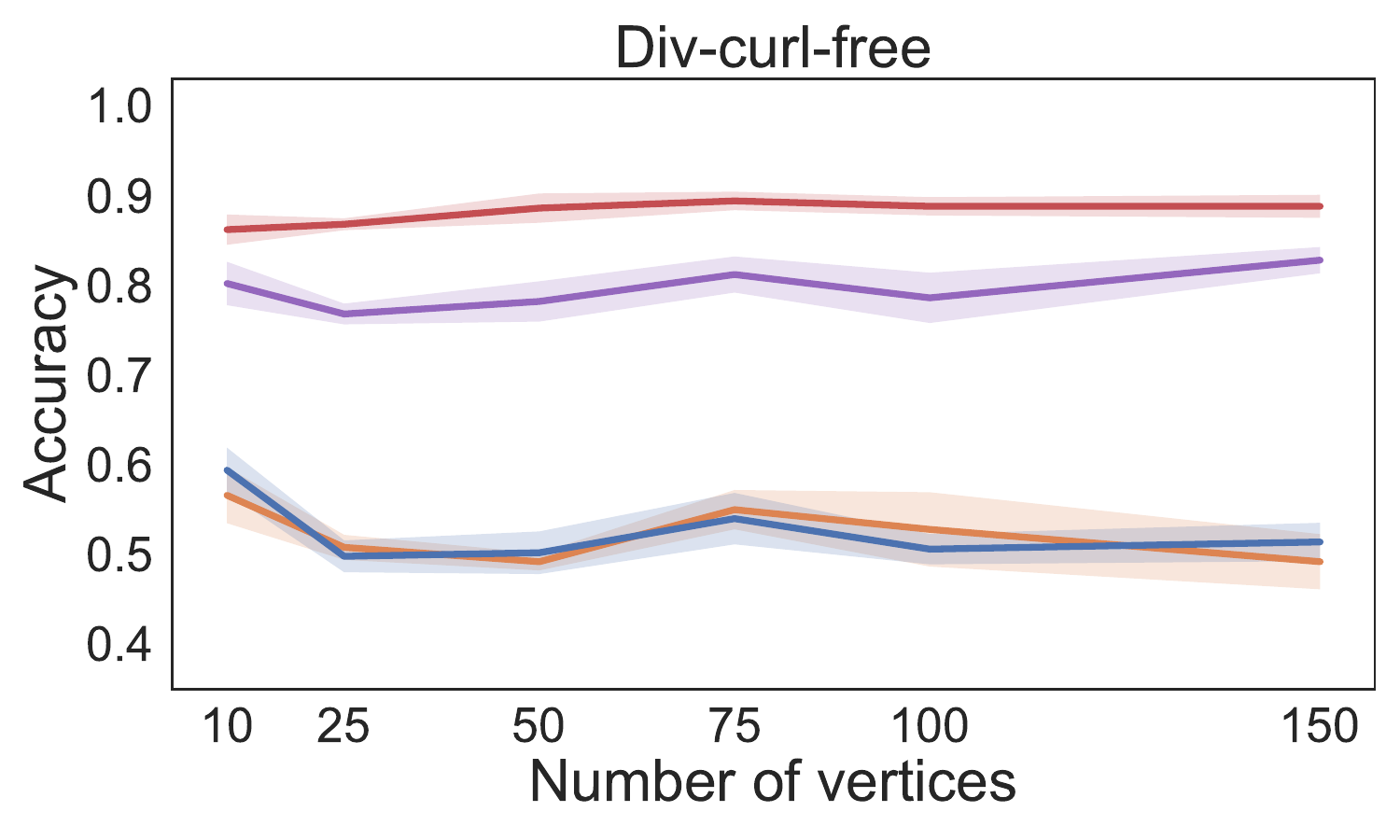}
        \label{fig:fig1}
    \end{minipage}
    \hfill
    \begin{minipage}{0.49\linewidth}
        \centering
        \includegraphics[width=\linewidth]{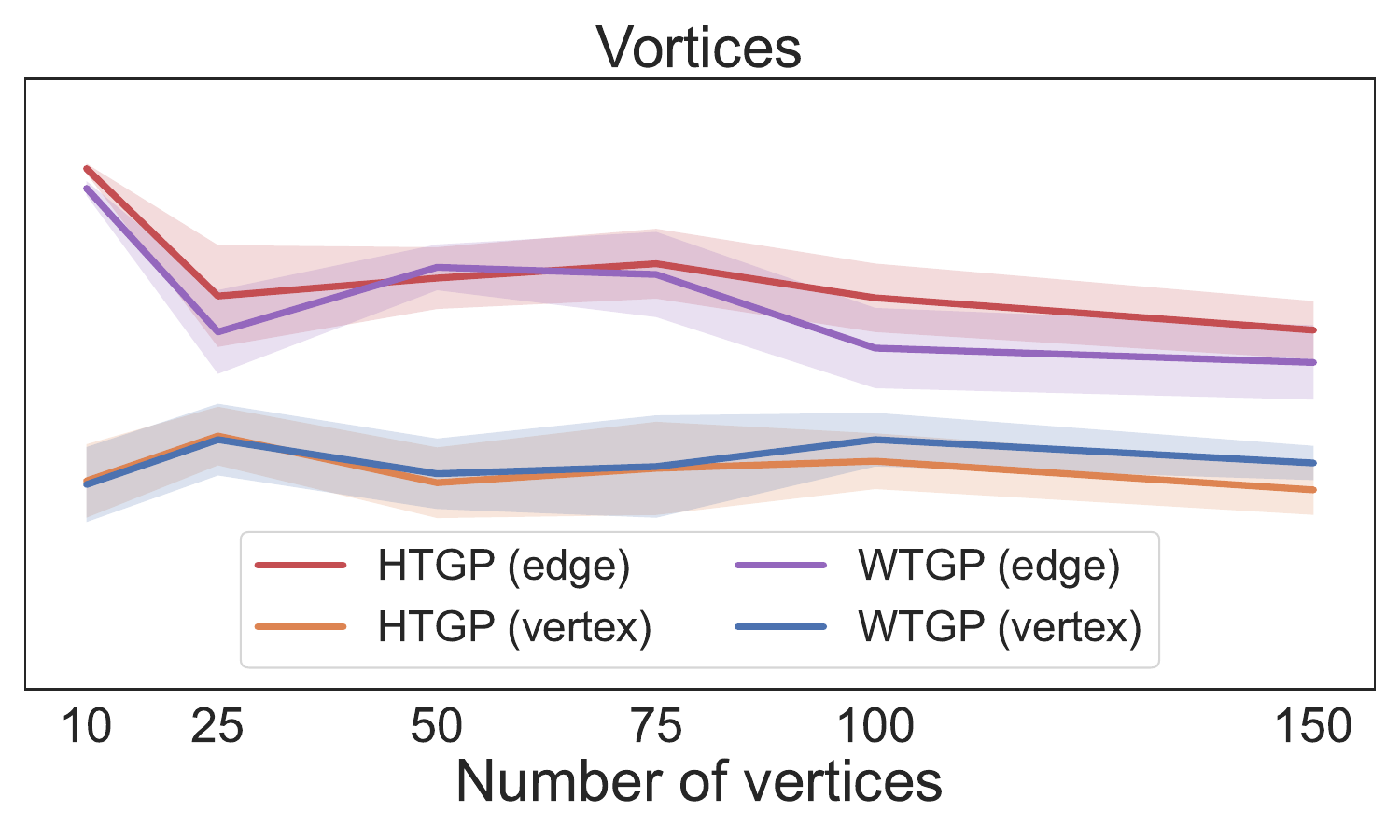}
        \label{fig:fig2}
    \end{minipage}
    \vspace{-\baselineskip}
    \caption{Accuracy of the GP models versus the mesh resolution (left: Div-curl-free, energy aggregation; right: Vortices, sum aggregation). Shaded regions indicate the standard error over five independently generated datasets of $100$ complexes.}
    \label{fig:divcurlfreevortices}
    \vspace{-3pt}
\end{figure}

Figure~\ref{fig:divcurlfreevortices} displays the accuracy of the GP models at different mesh resolutions. On both datasets, converting the edge-attributed complexes into line graphs with vertex attributes lowers the accuracy, highlighting the importance of preserving edge-level structure. On Div-curl-free, the line-graph models are close to chance level from $25$ vertices on ($0.49$ to $0.55$), whereas the edge models stay well above it; a likely reason is that line graphs lose the edge orientations, and with them the signed sums at vertices and around triangles that define divergence and curl. HTGP (edge) is the best-performing model on Div-curl-free at every resolution, ahead of WTGP (edge) by $0.06$ to $0.10$. On Vortices, HTGP (edge) and WTGP (edge) attain accuracies of $0.75$ to $0.93$ and $0.71$ to $0.91$, both well ahead of the line-graph models ($0.57$ to $0.63$). 

\subsection{TUDataset and MoleculeNet}\label{exp:TU}

The second series of datasets is taken from the standard molecular graph benchmarks TUDataset \citep{Morris2020} and MoleculeNet \citep{Wu2017}. The vertex and edge attributes are one-hot encodings of the atom and bond types (all atom and bond features for MoleculeNet). Unlike flows, bond attributes do not change sign under orientation reversal, so aggregations of their Hodgelet coefficients generally depend on the edge orientations, which follow the atom numbering; we therefore use the orientation-averaged sum and energy \eqref{eq:oavg}, which do not depend on these orientations. As for all graph benchmarks in this paper, each graph is lifted to a simplicial $2$-complex by filling every $3$-clique (see Appendix~\ref{appendix:sc}). MUTAG and AIDS are binary classification tasks, while FreeSolv and ESOL are regression tasks.
We summarise our results in Table~\ref{TUDatasets_MoleculeNet}. 
For MUTAG, the four GP models that use the edge attributes are the best-performing models, ahead of all GNN baselines (at most $0.802$) and of their vertex-only counterparts (by four to six points). For AIDS, all models are close to the accuracy ceiling. For FreeSolv and ESOL, WTGP (hybrid), WTGP (vertex), and HTGP (hybrid) are the three best-performing models and all outperform the GNN baselines. On both regression datasets, the bond attributes are less informative than the atom attributes, and the additive kernel compensates for this in the hybrid models, which are on par with or better than their vertex-only counterparts. On these molecular benchmarks, whose bond attributes are not flows, HTGP and WTGP perform similarly. On the flows of the vector fields and power grids, HTGP (edge) is generally ahead of WTGP (edge). Overall, the GP models are the best-performing models or on par with the best GNN on each dataset. On MUTAG, they are also better calibrated: every GP model has a lower NLPD and ECE than every GNN (e.g., an NLPD of $0.296$ for HTGP (hybrid), against $0.452$ for GIN, the best GNN; see Appendix~\ref{appendix:additional}).

\begin{table}[t]
\centering
\caption{Mean and standard error over five repetitions with different random splits. Colours indicate the {\color{BrickRed} best}, {\color{Purple} second-best}, and {\color{NavyBlue} third-best} result for each dataset.}
\label{TUDatasets_MoleculeNet}
{\small\begin{tabular}{lcccc}
\toprule
& \multicolumn{2}{c}{Accuracy ($\uparrow$)} & \multicolumn{2}{c}{Mean squared error ($\downarrow$)} \\
\cmidrule(lr){2-3}\cmidrule(lr){4-5}
 & MUTAG & AIDS ($\times 10^{-2}$) & FreeSolv & ESOL \\
\midrule
GCN (vertex) & $0.790 \pm 0.014$ & $99.60 \pm 0.04$ & $1.86 \pm 0.11$ & $0.47 \pm 0.01$ \\
GAT (hybrid) & $0.763 \pm 0.009$ & $99.59 \pm 0.02$ & $2.19 \pm 0.14$ & $\color{NavyBlue} \mathbf{0.45} \pm 0.01$ \\
GIN (hybrid) & $0.802 \pm 0.016$ & $99.54 \pm 0.02$ & $2.06 \pm 0.14$ & $1.02 \pm 0.08$ \\
Transformer (hybrid) & $0.789 \pm 0.016$ & $99.55 \pm 0.04$ & $1.99 \pm 0.23$ & $0.48 \pm 0.01$ \\
\midrule
WTGP (hybrid) & $\color{NavyBlue} \mathbf{0.863} \pm 0.010$ & $\color{NavyBlue} \mathbf{99.61} \pm 0.04$ & $\color{BrickRed} \mathbf{1.20} \pm 0.04$ & $\color{BrickRed} \mathbf{0.41} \pm 0.01$ \\
WTGP (vertex) & $0.819 \pm 0.006$ & $99.14 \pm 0.05$ & $\color{Purple} \mathbf{1.22} \pm 0.05$ & $\color{BrickRed} \mathbf{0.41} \pm 0.01$ \\
WTGP (edge) & $\color{BrickRed} \mathbf{0.878} \pm 0.007$ & $99.58 \pm 0.02$ & $9.22 \pm 0.16$ & $1.62 \pm 0.02$ \\
HTGP (hybrid) & $\color{BrickRed} \mathbf{0.878} \pm 0.009$ & $\color{BrickRed} \mathbf{99.64} \pm 0.01$ & $\color{NavyBlue} \mathbf{1.26} \pm 0.02$ & $\color{Purple} \mathbf{0.43} \pm 0.01$ \\
HTGP (vertex) & $0.815 \pm 0.006$ & $99.11 \pm 0.03$ & $1.32 \pm 0.02$ & $\color{NavyBlue} \mathbf{0.45} \pm 0.01$ \\
HTGP (edge) & $\color{Purple} \mathbf{0.870} \pm 0.003$ & $\color{Purple} \mathbf{99.63} \pm 0.03$ & $8.75 \pm 0.11$ & $1.59 \pm 0.01$ \\
\bottomrule
\end{tabular}
}
\vspace{-8pt}
\end{table}
\vspace{-7pt}
\subsection{PowerGraph}\label{sec:powergraph}
\vspace{-5pt}
In the last series of experiments, we consider prediction tasks in electricity networks using the ieee24 data from the PowerGraph benchmark \citep{Varbella2024}. We subsample this real-world dataset to the small-data regime (see Appendix~\ref{appendix: preprocessing}).
The graphs represent power grids at different pre-outage operating conditions, with electrical buses as vertices and transmission lines and transformers as edges. The vertex attributes are the three bus features provided by PowerGraph (the net active power, the net apparent power, and the voltage magnitude). The edge attributes consist of the active and reactive power flows, the branch reactance, and the line rating. Since all graphs share the same grid, a single edge orientation is used for every sample. The three tasks are to determine whether (binary classification) and to what extent (multiclass classification and regression) a power grid is stable. 
In Table~\ref{ieee24}, HTGP (hybrid) and HTGP (edge) are the two best-performing models on ieee24-binary, ahead of the best GNN, and the four GP models that use the edge attributes are also ahead of all GNN baselines on ieee24-multiclass. On ieee24-regression, HTGP (hybrid) and HTGP (edge) are again the two best-performing models. The edge attributes clearly improve performance: on both classification tasks, the GP models that use them outperform their vertex-only counterparts, by three to nine points on ieee24-binary and by fourteen to sixteen points on ieee24-multiclass. The Hodge decomposition also helps with these power flows on ieee24-binary and ieee24-regression, where HTGP (hybrid) and HTGP (edge) outperform their WTGP counterparts; on ieee24-multiclass, the HTGP and WTGP models are on par. HTGP (edge) also attains the lowest ECE on both classification tasks and the lowest NLPD, tied with WTGP (edge) on ieee24-multiclass (see Appendix~\ref{appendix:additional}).

\begin{table}[ht]

\centering
\caption{Mean and standard error over five independent subsamples. Colours indicate the {\color{BrickRed} best}, {\color{Purple} second-best}, and {\color{NavyBlue} third-best} result for each dataset.}
\label{ieee24}
{\small\begin{tabular}{lccc}
\toprule
& \multicolumn{2}{c}{Accuracy ($\uparrow$)} & Mean squared error ($\downarrow$) \\
\cmidrule(lr){2-3}\cmidrule(lr){4-4}
 & ieee24-binary & ieee24-multiclass & ieee24-regression ($\times 10^{-4}$) \\
\midrule
GCN (vertex) & $0.767 \pm 0.010$ & $0.689 \pm 0.026$ & $4.21 \pm 0.31$ \\
GAT (hybrid) & $0.859 \pm 0.010$ & $0.872 \pm 0.009$ & $3.77 \pm 0.26$ \\
GIN (hybrid) & $\color{NavyBlue} \mathbf{0.885} \pm 0.014$ & $0.876 \pm 0.010$ & $2.58 \pm 0.19$ \\
Transformer (hybrid) & $0.879 \pm 0.013$ & $0.889 \pm 0.004$ & $3.48 \pm 0.32$ \\
\midrule
WTGP (hybrid) & $0.879 \pm 0.013$ & $0.906 \pm 0.004$ & $\color{NavyBlue} \mathbf{2.19} \pm 0.10$ \\
WTGP (vertex) & $0.811 \pm 0.006$ & $0.766 \pm 0.011$ & $3.05 \pm 0.24$ \\
WTGP (edge) & $0.843 \pm 0.025$ & $\color{BrickRed} \mathbf{0.921} \pm 0.003$ & $4.79 \pm 0.81$ \\
HTGP (hybrid) & $\color{BrickRed} \mathbf{0.906} \pm 0.010$ & $\color{NavyBlue} \mathbf{0.911} \pm 0.008$ & $\color{BrickRed} \mathbf{1.61} \pm 0.20$ \\
HTGP (vertex) & $0.813 \pm 0.009$ & $0.759 \pm 0.017$ & $3.16 \pm 0.21$ \\
HTGP (edge) & $\color{Purple} \mathbf{0.902} \pm 0.007$ & $\color{Purple} \mathbf{0.919} \pm 0.007$ & $\color{Purple} \mathbf{1.84} \pm 0.24$ \\
\bottomrule
\end{tabular}
}
\vspace{-15pt}
\end{table}

\section{Conclusion}
\vspace{-5pt}
In this paper, we propose a representation of attributed SCs and graphs for prediction tasks, which, combined with GPs, gives a robust and interpretable method that achieves competitive performance and well-calibrated predictions across a variety of datasets. Our Hodgelet representations leverage both the wavelet transform of the attributes, to capture multi-scale spectral and spatial information, and the Hodge decomposition, to separate the signal into exact, co-exact, and harmonic components for finer control. In our experiments, the GP models match or outperform the GNN baselines on the molecular and power-grid benchmarks; keeping the edge attributes on the edges, rather than moving them to a line graph, substantially improves accuracy on vector fields; and separating the Hodge components helps most on Div-curl-free and the power grids. On the orientability of triangulated surfaces, a purely topological task, the GP models are almost perfect, well ahead of the neural baselines (see Appendix~\ref{appendix:mantra}). In future work, we intend to incorporate the geometry of the SC, which has been shown to be useful for molecular prediction tasks \citep{fang2022geometry}, for example through its curvature \citep{Ni2019}, and to adapt our approach to prediction tasks at the level of vertices or edges.

\begin{ack}
MA is supported by a Mathematical Sciences Doctoral Training Partnership held by Prof.~Helen Wilson, funded by the Engineering and Physical Sciences Research Council (EPSRC), under Project Reference EP/W523835/1. ST is supported by a Department of Defense Vannevar Bush Faculty Fellowship held by Prof.~Andrew Stuart, and by the SciAI Center, funded by the Office of Naval Research (ONR), under Grant Number N00014-23-1-2729.
\end{ack}

\bibliographystyle{plainnat}
\bibliography{references}

@inproceedings{Opolka2023,
    title={Graph Classification {Gaussian} Processes via Spectral Features},
    author={Felix Opolka and Yin-Cong Zhi and Pietro Liò and Xiaowen Dong},
    booktitle={Conference on Uncertainty in Artificial Intelligence},
    year={2023}}

@inproceedings{bodnar2021weisfeiler,
  title={Weisfeiler and {L}ehman Go Topological:
         Message Passing Simplicial Networks},
  author={Bodnar, Cristian and Frasca, Fabrizio
          and Wang, Yuguang and Otter, Nina
          and Montufar, Guido F and Li{\`o}, Pietro
          and Bronstein, Michael},
  booktitle={Proceedings of the 38th International
             Conference on Machine Learning},
  pages={1026--1037},
  volume={139},
  series={Proceedings of Machine Learning Research},
  year={2021},
  publisher={PMLR}
}

@inproceedings{Battiloro2023,
    title={Topological {Slepians}: Maximally Localized Representations of Signals Over Simplicial Complexes},
    author={Claudio Battiloro and Paolo Di Lorenzo and Sergio Barbarossa},
    booktitle={IEEE International Conference on Acoustics, Speech and Signal Processing },
    year={2023}}

@article{Hammond2011,
    author={Hammond, David K and Vandergheynst, Pierre and Gribonval, Rémi},
    journal={Applied and Computational Harmonic Analysis}, 
    title={Wavelets on graphs via spectral graph theory}, 
    year={2011}}

@inproceedings{Yang2024,
    title={{Hodge}-Compositional Edge {Gaussian} Processes},
    author={Maosheng Yang and Viacheslav Borovitskiy and Elvin Isufi},
    booktitle={International Conference on Artificial Intelligence and Statistics},
    year={2024}}

@article{Ortega2018,
    author={Antonio Ortega and Pascal Frossard and Jelena Kova{\v{c}}evi{\'c} and José MF Moura and Pierre Vandergheynst},
    journal={Proceedings of the IEEE}, 
    title={Graph signal processing: Overview, challenges, and applications}, 
    year={2018}}

@article{Lim2020,
    author ={Lek-Heng Lim},
    journal ={SIAM Review},
    title = {{Hodge} {Laplacians} on Graphs},
    year = {2020}}

@book{Rasmussen2005,
    author ={Carl Edward Rasmussen and Christopher K I Williams},
    title = {{Gaussian} processes for machine learning},
    publisher = {MIT Press},
    year = {2005}}

@inproceedings{Feng2019,
    title={Hypergraph Neural Networks},
    author = {Yifan Feng and Haoxuan You and Zizhao Zhang and Rongrong Ji and Yue Gao},
    booktitle={AAAI Conference on Artificial Intelligence},
    year={2019}}

@article{Barbarossa2020,
    author ={Sergio Barbarossa and Stefania Sardellitti},
    journal={IEEE Transactions on Signal Processing}, 
    title={Topological signal processing over simplicial complexes}, 
    year={2020}}

@inproceedings{Alain2024,
    author = {Mathieu Alain and So Takao and Brooks Paige and Marc P Deisenroth},
    title = {{Gaussian} Processes on Cellular Complexes},
    booktitle = {International Conference on Machine Learning},
    year = {2024}
}

@inproceedings{Hu2020,
    title={{S}trategies for {P}re-training {G}raph {N}eural {N}etworks},
    author={Weihua Hu and Bowen Liu and Joseph Gomes and Marinka Zitnik and Percy Liang and Vijay Pande and Jure Leskovec},
    booktitle={International Conference on Learning Representations},
    year={2020}}

@inproceedings{Shi2021,
    title={{M}asked {L}abel {P}rediction: {U}nified {M}essage {P}assing {M}odel for {S}emi-{S}upervised
{C}lassification},
    author={Yunsheng Shi and Zhengjie Huang and Shikun Feng and Hui Zhong and Wenjing Wang and Yu Sun},
    booktitle={International Joint Conference on Artificial Intelligence},
    year={2021}}

@inproceedings{Velickovic2018,
    title={Graph {A}ttention {N}etworks},
    author={Petar Veli{\v{c}}kovi{\'c} and Guillem Cucurull and Arantxa Casanova and Adriana Romero and Pietro Liò and Yoshua Bengio},
    booktitle={International Conference on Learning Representations},
    year={2018}}

@inproceedings{Kipf2017,
    title={Semi-supervised {C}lassification with {G}raph {C}onvolutional {N}etworks},
    author={Thomas Kipf and Max Welling},
    booktitle={International Conference on Learning Representations},
    year={2017}}

@inproceedings{Papamarkou2024,
    author = {Theodore Papamarkou and Tolga Birdal and Michael Bronstein and Gunnar Carlsson and Justin Curry and Yue Gao and Mustafa Hajij and Roland Kwitt and Pietro Liò and Paolo Di Lorenzo and Vasileios Maroulas and Nina Miolane and Farzana Nasrin and Karthikeyan Natesan Ramamurthy and Bastian Rieck and Simone Scardapane and Michael T Schaub and Petar Veli{\v{c}}kovi{\'c} and Bei Wang and Yusu Wang and Guo-Wei Wei and Ghada Zamzmi},
    title = {Position: Topological Deep Learning is the New Frontier for Relational Learning},
    booktitle = {International Conference on Machine Learning},
    year = {2024}
}

@article{Reiser2022,
    author ={Reiser, Patrick and Neubert, Marlen and Eberhard, André and Torresi, Luca and Zhou, Chen and Shao, Chen and  Metni, Houssam and van Hoesel, Clint  and Schopmans, Henrik and Sommer, Timo and Friederich, Pascal},
    journal={Communications Materials}, 
    title={Graph neural networks for materials science and chemistry}, 
    year={2022}}

@inproceedings{Defferrard2016,
    title={Convolutional Neural Networks on Graphs with Fast Localized Spectral Filtering},
    author={Defferrard, Michaël and Bresson, Xavier and Vandergheynst, Pierre},
    booktitle={Conference on Neural Information Processing Systems},
    year={2016}}

@inproceedings{li2018deeper,
  title={Deeper Insights into Graph Convolutional Networks for Semi-Supervised Learning},
  author={Li, Qimai and Han, Zhichao and Wu, Xiao-Ming},
  booktitle={Proceedings of the AAAI Conference on Artificial Intelligence},
  year={2018}
}

@article{kriege2020survey,
  title={A Survey on Graph Kernels},
  author={Kriege, Nils M. and Johansson, Fredrik D. and Morris, Christopher},
  journal={Applied Network Science},
  year={2020},
  publisher={Springer}
}

@inproceedings{rong2020dropedge,
  title={{DropEdge}: Towards Deep Graph Convolutional Networks on Node Classification},
  author={Rong, Yu and Huang, Wenbing and Xu, Tingyang and Huang, Junzhou},
  booktitle={International Conference on Learning Representations},
  year={2020}
}

@inproceedings{Roddenberry2022,
    title={Hodgelets: Localized Spectral Representations of Flows on Simplicial Complexes},
    author={Roddenberry, T. Mitchell and Frantzen, Florian and Schaub, Michael T. and Segarra, Santiago},
    booktitle={International Conference on Acoustics, Speech, and Signal Processing},
    year={2022}}

@inproceedings{Morris2020,
    author = {Christopher Morris and Nils M. Kriege and Franka Bause and Kristian Kersting and Petra Mutzel and Marion Neumann},
    title = {{TUDataset}: A collection of benchmark datasets for learning with graphs},
    booktitle = {ICML 2020 Workshop on Graph Representation Learning and Beyond},
    year = {2020}
}

@inproceedings{Varbella2024,
    author = {Anna Varbella and Kenza Amara and Blazhe Gjorgiev and Mennatallah El-Assady and Giovanni Sansavini},
    title = {{PowerGraph}: A power grid benchmark dataset for graph neural networks},
    booktitle = {NeurIPS 2024 Track on Datasets and Benchmarks},
    year = {2024}
}

@article{rahimi2007random,
  title={Random features for large-scale kernel machines},
  author={Rahimi, Ali and Recht, Benjamin},
  journal={Advances in neural information processing systems},
  volume={20},
  year={2007}
}

@article{Opper2009,
  title={The Variational {Gaussian} Approximation Revisited},
  author={Manfred Opper and Cédric Archambeau},
  journal={Neural Computation},
  year={2009}
}

@article{Isufi2025,
  title={Topological signal processing and learning: Recent advances and future challenges},
  author={Elvin Isufi and Geert Leus and Baltasar Beferull-Lozano and Sergio Barbarossa and Paolo Di Lorenzo},
  journal={Signal Processing},
  year={2025}
}

@article{Wu2017,
      title={{MoleculeNet}: A Benchmark for Molecular Machine Learning}, 
      author={Zhenqin Wu and Bharath Ramsundar and Evan N. Feinberg and Joseph Gomes and Caleb Geniesse and Aneesh S. Pappu and Karl Leswing and Vijay Pande},
      year={2018},
      journal={Chemical Science},
      volume={9},
      number={2},
      pages={513--530}
}

@inproceedings{Ballester2024,
      title={{MANTRA}: The Manifold Triangulations Assemblage},
      author={Rubén Ballester and Ernst Röell and Bin Schmid, Daniel and Mathieu Alain and Sergio Escalera and Carles Casacuberta and Bastian Rieck},
      booktitle={International Conference on Learning Representations},
      year={2025}
}

@article{Borgwardt2020,
  title={Graph Kernels: State-of-the-Art and Future Challenges},
  author={Karsten Borgwardt and Elisabetta Ghisu and Felipe Llinares-López and Leslie O'Bray and Bastian Rieck},
  journal={Foundations and Trends in Machine Learning},
  year={2020}
}

@article{Titsias2009,
  title={Variational learning of inducing variables in sparse {Gaussian} processes},
  author={Michalis Titsias},
  journal={International Conference on Artificial intelligence and Statistics},
  year={2009}
}

@Inbook{Desbrun2008,
author="Desbrun, Mathieu
and Kanso, Eva
and Tong, Yiying",
title="Discrete Differential Forms for Computational Modeling",
bookTitle="Discrete Differential Geometry",
year="2008",
publisher="Birkh{\"a}user Basel",
pages="287--324",
}

@article{Ni2019,
  title={Community Detection on Networks with {Ricci} Flow},
  author={Chien-Chun Ni and Yu-Yao Lin and Feng Luo and Jie Gao},
  journal={Scientific Reports},
  year={2019}
}

@article{Baccini2022,
  title={Weighted simplicial complexes and their representation power of higher-order network data and topology},
  author={Federica Baccini and Filippo Geraci and Ginestra Bianconi},
  journal={Physical Review E},
  year={2022}
}

@book{Bronstein2021,
    author = {Michael Bronstein and Joan Bruna and Taco Cohen and Petar Veli{\v{c}}kovi{\'c}},
    title = {Geometric Deep Learning: Grids, Groups, Graphs, Geodesics, and Gauges},
    year = {2021},
    publisher={arXiv}
}

@inproceedings{pytorch2019,
 author = {Paszke, Adam and Gross, Sam and Massa, Francisco and Lerer, Adam and Bradbury, James and Chanan, Gregory and Killeen, Trevor and Lin, Zeming and Gimelshein, Natalia and Antiga, Luca and Desmaison, Alban and Kopf, Andreas and Yang, Edward and DeVito, Zachary and Raison, Martin and Tejani, Alykhan and Chilamkurthy, Sasank and Steiner, Benoit and Fang, Lu and Bai, Junjie and Chintala, Soumith},
 booktitle = {Advances in Neural Information Processing Systems},
 title = {{PyTorch}: An Imperative Style, High-Performance Deep Learning Library},
 year = {2019}
}

@inproceedings{pygeometric2019,
  title={Fast Graph Representation Learning with {PyTorch Geometric}},
  author={Fey, Matthias and Lenssen, Jan E.},
  booktitle={ICLR Workshop on Representation Learning on Graphs and Manifolds},
  year={2019},
}

@inproceedings{gpytorch2018,
 author = {Gardner, Jacob and Pleiss, Geoff and Weinberger, Kilian Q and Bindel, David and Wilson, Andrew G},
 booktitle = {Advances in Neural Information Processing Systems},
 title = {{GPyTorch}: Blackbox Matrix-Matrix {Gaussian} Process Inference with {GPU} Acceleration},
 year = {2018}
}

@inproceedings{
Alain2024b,
title={{G}raph {C}lassification {G}aussian {P}rocesses
via {H}odgelet {S}pectral {F}eatures},
author={Mathieu Alain and So Takao and Bastian Rieck and Xiaowen Dong and Emmanuel Noutahi},
booktitle={{A}dvances in {N}eural {I}nformation {P}rocessing {S}ystems (NeurIPS) 2024 Workshop Workshop on Bayesian Decision-making and Uncertainty (BDU)},
year={2024},
url={https://arxiv.org/pdf/2410.10546}
}

@article{luo2020parameterized,
  title={Parameterized explainer for graph neural network},
  author={Luo, Dongsheng and Cheng, Wei and Xu, Dongkuan and Yu, Wenchao and Zong, Bo and Chen, Haifeng and Zhang, Xiang},
  journal={Advances in neural information processing systems},
  year={2020}
}

@article{ying2019gnnexplainer,
  title={{GNNExplainer}: Generating explanations for graph neural networks},
  author={Ying, Zhitao and Bourgeois, Dylan and You, Jiaxuan and Zitnik, Marinka and Leskovec, Jure},
  journal={Advances in neural information processing systems},
  volume={32},
  year={2019}
}

@inproceedings{gilmer2017neural,
  title={Neural message passing for quantum chemistry},
  author={Gilmer, Justin and Schoenholz, Samuel S and Riley, Patrick F and Vinyals, Oriol and Dahl, George E},
  booktitle={International conference on machine learning},
  pages={1263--1272},
  year={2017},
  organization={PMLR}
}

@inproceedings{hensman2013gaussian,
  title={{Gaussian} processes for big data},
  author={Hensman, James and Fusi, Nicolo and Lawrence, Neil D},
  booktitle={Conference on Uncertainty in Artificial Intelligence},
  year={2013}
}

@article{fang2022geometry,
  title={Geometry-enhanced molecular representation learning for property prediction},
  author={Fang, Xiaomin and Liu, Lihang and Lei, Jieqiong and He, Donglong and Zhang, Shanzhuo and Zhou, Jingbo and Wang, Fan and Wu, Hua and Wang, Haifeng},
  journal={Nature Machine Intelligence},
  volume={4},
  number={2},
  pages={127--134},
  year={2022},
  publisher={Nature Publishing Group UK London}
}

@article{beale1985high,
  title={High order accurate vortex methods with explicit velocity kernels},
  author={Beale, J Thomas and Majda, Andrew},
  journal={Journal of Computational Physics},
  volume={58},
  number={2},
  pages={188--208},
  year={1985},
  publisher={Elsevier}
}

@book{newton2002n,
  title={The {$N$}-Vortex Problem: Analytical Techniques},
  author={Newton, Paul K},
  series={Applied Mathematical Sciences},
  publisher={Springer},
  address={New York},
  year={2001}
}

@article{pinder2021gaussian,
  title={{Gaussian} processes on hypergraphs},
  author={Pinder, Thomas and Turnbull, Kathryn and Nemeth, Christopher and Leslie, David},
  journal={arXiv preprint arXiv:2106.01982},
  year={2021}
}

@book{Petersen2016,
  author    = {Peter Petersen},
  title     = {Riemannian Geometry},
  edition   = {3},
  publisher = {Springer},
  year      = {2016},
  series    = {Graduate Texts in Mathematics},
  volume    = {171}
}

@inproceedings{Rousseau2015,
  author    = {Fran{\c{c}}ois Rousseau and Emmanouil Kiagias and Michalis Vazirgiannis},
  title     = {Text Categorization as a Graph Classification Problem},
  booktitle = {Proceedings of the 53rd Annual Meeting of the Association for Computational Linguistics},
  pages     = {1702--1712},
  year      = {2015},
  doi       = {10.3115/v1/P15-1164}
}

@article{maloddognin2019functional,
  title={Functional geometry of protein interactomes},
  author={Malod-Dognin, No{\"e}l and Pr{\v{z}}ulj, Nata{\v{s}}a},
  journal={Bioinformatics},
  volume={35},
  number={19},
  pages={3727--3734},
  year={2019},
  publisher={Oxford University Press},
  doi={10.1093/bioinformatics/btz146}
}

@inproceedings{ebli2020simplicial,
  title={Simplicial Neural Networks},
  author={Stefania Ebli and Micha{\"e}l Defferrard and Gard Spreemann},
  booktitle={NeurIPS 2020 Workshop on Topological Data Analysis and Beyond},
  year={2020}
}

@article{yang2022simplicial,
  title={Simplicial Convolutional Filters},
  author={Maosheng Yang and Elvin Isufi and Michael T. Schaub and Geert Leus},
  journal={IEEE Transactions on Signal Processing},
  volume={70},
  pages={4633--4648},
  year={2022}
}

@article{shervashidze2011weisfeiler,
  title={{W}eisfeiler--{L}ehman Graph Kernels},
  author={Nino Shervashidze and Pascal Schweitzer and Erik Jan van Leeuwen and Kurt Mehlhorn and Karsten M. Borgwardt},
  journal={Journal of Machine Learning Research},
  volume={12},
  pages={2539--2561},
  year={2011}
}

@inproceedings{fuchsgruber2025graph,
  title={Graph Neural Networks for Edge Signals: Orientation Equivariance and Invariance},
  author={Dominik Fuchsgruber and Tim Postuvan and Stephan G{\"u}nnemann and Simon Geisler},
  booktitle={International Conference on Learning Representations},
  year={2025}
}

@inproceedings{xu2019powerful,
  title={How Powerful are Graph Neural Networks?},
  author={Xu, Keyulu and Hu, Weihua and Leskovec, Jure and Jegelka, Stefanie},
  booktitle={International Conference on Learning Representations},
  year={2019}}

@article{Garreau2017,
  title = {Large sample analysis of the median heuristic},
  author = {Garreau, Damien and Jitkrittum, Wittawat and Kanagawa, Motonobu},
  journal = {arXiv preprint arXiv:1707.07269},
  year = {2017}
}

@inproceedings{guo2017calibration,
    author = {Chuan Guo and Geoff Pleiss and Yu Sun and Kilian Q Weinberger},
    title = {On Calibration of Modern Neural Networks},
    booktitle = {International Conference on Machine Learning},
    year = {2017}
}

@inproceedings{kuleshov2018accurate,
    author = {Volodymyr Kuleshov and Nathan Fenner and Stefano Ermon},
    title = {Accurate Uncertainties for Deep Learning Using Calibrated Regression},
    booktitle = {International Conference on Machine Learning},
    year = {2018}
}

\clearpage
\appendix

\section{Simplicial complexes}\label{appendix:sc}

In this appendix, we provide details on simplicial complexes, which are the key objects of this work. A graph consists of a set $V$ of vertices and a set $E$ of edges, where each edge is a pair $e=\{v_0,v_1\}$ of distinct vertices. 
Graphs can carry signals on their vertices and edges, resulting in a structure that is richer than tabular data. Common examples are social networks, electrical grids (see Section~\ref{sec:powergraph}), citation networks, traffic maps, collaboration networks, chemical reaction networks, vector fields (see Section~\ref{exp:vectorfield}), and molecules (see Section~\ref{exp:TU}). However, a limitation of graphs is that they can only encode \emph{dyadic interactions} between vertices through their edges. Multiple recent works have sought to overcome this limitation by using more general structures, such as simplicial complexes, cellular complexes, and hypergraphs \citep{Feng2019, Baccini2022, Roddenberry2022,  Alain2024, Yang2024, Papamarkou2024}. 

\paragraph{Simplicial complexes.}
Simplicial complexes are a natural extension of graphs, built from \emph{$k$-simplices}. A $k$-simplex is a set of $k+1$ vertices: vertices are $0$-simplices, edges are $1$-simplices, and a $2$-simplex, also called a triangular face or a triangle, is formed by three pairwise adjacent vertices and describes a \emph{triadic interaction} between them.

A \emph{face} of a $k$-simplex is a $(k-1)$-simplex formed by a subset of $k$ of the vertices defining the $k$-simplex.
Then, as in Section~\ref{subsection:SC}, a \emph{simplicial $K$-complex} is a collection of simplex sets $S_0, \ldots, S_K$, where $S_k$ is a set of $k$-simplices, such that the faces of every simplex are also simplices of the complex. In other words, the union $S_0 \cup \cdots \cup S_K$ is a set of sets, each with at most $K+1$ elements, that is closed under the operation of taking non-empty subsets. For example, a graph $G = (V, E)$ is a simplicial $1$-complex with $S_0 = \{\{v\} : v \in V\}$ and $S_1 = E$, since both vertices of every edge are in $V$.

In this work, we are primarily concerned with simplicial $2$-complexes, which consist of the sets $S_0$, $S_1$, and $S_2$ of vertices, edges, and triangular faces, respectively. By our definition above, these sets are required to satisfy the following closure conditions:
\begin{enumerate}
    \item all vertices that form an edge in $S_1$ must be in $S_0$, and
    \item all edges that form a triangle in $S_2$ must be in $S_1$. 
\end{enumerate}
Since an edge is a set of two distinct vertices and is determined by them, there are no self-loops or multi-edges. 
Finally, as in Section~\ref{subsection:SC}, the numbers of vertices, edges, and triangles are denoted by $N_0$, $N_1$, and $N_2$, respectively. 

\paragraph{Orientation.} An orientation of a simplicial complex assigns to each simplex $\{v_0, \ldots, v_k\} \in S_k$, for $0 \leq k \leq K$, an ordering of its vertices, where orderings that differ by an even permutation are considered equivalent. We denote by $[(v_0, \ldots, v_k)]$ the equivalence class of the ordering $(v_0, \ldots, v_k)$, and say that the simplex is oriented as $(v_0, \ldots, v_k)$. An edge $\{v_0, v_1\}$ thus has two orientations, the directions $(v_0, v_1)$ and $(v_1, v_0)$, and a triangle $\{v_0, v_1, v_2\}$ has the two orientations
\begin{align*}
    &[(v_0, v_1, v_2)] = \{(v_0, v_1, v_2), (v_1, v_2, v_0), (v_2, v_0, v_1)\}, \\
    &[(v_2, v_1, v_0)] = \{(v_2, v_1, v_0), (v_1, v_0, v_2), (v_0, v_2, v_1)\},
\end{align*}
which correspond to the two directions of traversing the triangle, clockwise and anticlockwise. An orientation is needed to define edge flows and the incidence matrices below, but its choice is arbitrary.

Given an orientation, the entry $(\bm B_k)_{ij}$ of the incidence matrix is $\pm1$ if the $i$-th $(k-1)$-simplex is a face of the $j$-th $k$-simplex, and $0$ otherwise. The sign is positive if the orientation of the face agrees with the orientation that the $k$-simplex induces on it, and negative otherwise. Here, $[(v_0,\ldots,v_k)]$ induces on the face without $v_l$ the orientation $[(v_0,\ldots,v_{l-1},v_{l+1},\ldots,v_k)]$ if $l$ is even, and the opposite orientation if $l$ is odd. For example, the column of $\bm B_1$ of an edge oriented as $(v_0,v_1)$ has the entries $-1$ at $v_0$ and $1$ at $v_1$, and the column of $\bm B_2$ of a triangle oriented as $(v_0,v_1,v_2)$ has the entries $1$, $-1$, and $1$ at the edges oriented as $(v_1,v_2)$, $(v_0,v_2)$, and $(v_0,v_1)$, respectively. With this convention, $\bm B_k\bm B_{k+1}=\bm 0$, and reversing the orientation of a $k$-simplex flips the sign of its column of $\bm B_k$ and of its row of $\bm B_{k+1}$. In our experiments, the orientation is induced by the vertex numbering: every simplex is oriented by increasing vertex index.

\paragraph{Orientation-averaged aggregations.} Here, we derive the orientation-averaged sum and energy \eqref{eq:oavg}, which are introduced in this work, and show that they are invariant under changes of orientation. When the attributes are simplicial cochains, such as edge flows, the arbitrary choice of orientation is not an issue for aggregations that are invariant under sign changes, such as the energy, since the coefficients then change sign together with the orientation. However, many attributes, such as molecular bond types, do not change sign when the orientation of a simplex is reversed. Applying wavelet filters of the signed Hodge Laplacian to such attributes makes the Hodgelet coefficients, and therefore their aggregation, depend on the chosen orientation. \citet{fuchsgruber2025graph} identify the same problem for graph neural networks acting on edge signals, and process such orientation-invariant signals with unsigned incidence matrices instead of signed ones. Unsigned incidence matrices, however, do not satisfy $\bm B_k\bm B_{k+1}=\bm 0$, so that the resulting Laplacian no longer admits a Hodge decomposition into orthogonal exact, co-exact, and harmonic subspaces. Since this decomposition is central to our approach, we instead keep the signed Hodge Laplacian and average the aggregation over all possible orientations.

More precisely, consider the Hodgelet transform $\mathbf W_{\!P_{k\bullet}}$ of any one Hodge component, $\bullet\in\{e,c,h\}$, defined in Section~\ref{sec:RH_transform}, and note that it is symmetric. Recall from Section~\ref{sec:representations} that a change of orientation is given by a diagonal sign matrix $\bm S=\mathrm{diag}(s_1,\ldots,s_{N_k})$. To average over all orientations, we take the signs $s_i\in\{-1,1\}$ to be independent, each equal to $1$ or $-1$ with probability $1/2$. In particular, $\bb E[s_is_j]=\delta_{ij}$, where $\delta_{ij}$ denotes the Kronecker delta. The orientation-averaged sum is then given by
\begin{equation}
\bb E_{\bm S}\big[\bm 1^\top\bm S\mathbf W_{\!P_{k\bullet}}\bm S\bm x_k\big]
=\sum_{i,j}\bb E[s_is_j]\,(\mathbf W_{\!P_{k\bullet}})_{ij}\,(\bm x_k)_j
=\sum_{i}(\mathbf W_{\!P_{k\bullet}})_{ii}\,(\bm x_k)_i.
\end{equation}
Similarly, using $\bm S^2=\bm I$ and the symmetry of $\mathbf W_{\!P_{k\bullet}}$, the orientation-averaged energy is given by
\begin{equation}
\begin{split}
\bb E_{\bm S}\big[\|\bm S\mathbf W_{\!P_{k\bullet}}\bm S\bm x_k\|^2\big]
&=\bb E_{\bm S}\big[\bm x_k^\top\bm S\mathbf W_{\!P_{k\bullet}}^2\bm S\bm x_k\big]\\
&=\sum_{i,j}\bb E[s_is_j]\,(\mathbf W_{\!P_{k\bullet}}^2)_{ij}\,(\bm x_k)_i(\bm x_k)_j
=\sum_{i}(\mathbf W_{\!P_{k\bullet}}^2)_{ii}\,(\bm x_k)_i^2.
\end{split}
\end{equation}
In other words, averaging over all orientations removes every term that involves two different simplices and leaves only the diagonals of $\mathbf W_{\!P_{k\bullet}}$ and of its square.

We now show that both quantities are invariant. Consider a fixed change of orientation of the $k$-simplices, given by a diagonal sign matrix $\bm D\in\R^{N_k\times N_k}$ (we write $\bm D$ to distinguish it from the random $\bm S$ above), whose $i$-th diagonal entry is $-1$ if the orientation of the $i$-th $k$-simplex is reversed, and $1$ otherwise. This change replaces $\bm B_k$ by $\bm B_k\bm D$ and $\bm B_{k+1}$ by $\bm D\bm B_{k+1}$, while changes of orientation of the $(k-1)$- and $(k+1)$-simplices cancel out in $\bm L_k$. As a result, $\bm L_k^{\mathrm{low}}$ and $\bm L_k^{\mathrm{up}}$ become $\bm D\bm L_k^{\mathrm{low}}\bm D$ and $\bm D\bm L_k^{\mathrm{up}}\bm D$, which have the same eigenvalues, with eigenvectors multiplied by $\bm D$, so $\mathbf W_{\!P_{k\bullet}}$ becomes $\bm D\mathbf W_{\!P_{k\bullet}}\bm D$, whose square is $\bm D\mathbf W_{\!P_{k\bullet}}^2\bm D$. Multiplying a matrix by $\bm D$ on both sides leaves its diagonal unchanged, so the diagonals of $\mathbf W_{\!P_{k\bullet}}$ and of its square are preserved; the attributes $\bm x_k$, which are not cochains, do not change either. Hence, the orientation-averaged sum and energy are invariant under changes of orientation. Similarly, a relabelling of the simplices permutes the diagonals and the attributes in the same way, which leaves both sums unchanged. Since the orientation is induced by the vertex numbering, a relabelling of the vertices combines such a permutation with a change of orientation, and both aggregations are also invariant under vertex relabellings.

\paragraph{Lifting graphs to simplicial complexes.} The graph benchmarks (TUDataset, MoleculeNet, and ieee24) are lifted to simplicial $2$-complexes: the vertices and edges of a graph are its $0$- and $1$-simplices, and every $3$-clique, that is, every set of three pairwise adjacent vertices, is added as a $2$-simplex. The lift is deterministic and commutes with vertex relabellings: isomorphic graphs yield isomorphic complexes. The triangles are found by intersecting the neighbourhoods of the two endpoints of each edge, in time at most $\cal O\big(\sum_{v\in V}\deg(v)^2\big)$. The graphs carry no attributes on the triangles, so the triangles enter our models only through $\bm B_2$. The meshes and MANTRA are already simplicial complexes and keep their native triangles. Table~\ref{tab:datasets} reports the resulting numbers of triangles, and Table~\ref{tab:triangles} the effect of removing them.

\paragraph{Cellular complexes and hypergraphs.} Simplicial complexes can be further generalised to cellular complexes, to which most of the constructions above extend \citep{Alain2024}. Hypergraphs, in which an edge can join any number of vertices, could similarly be handled using Laplacians specialised for hypergraphs \citep{pinder2021gaussian}.

\section{Variational Gaussian processes}
\label{appendix:variational_gp}

For regression with Gaussian observation noise, the conjugacy between
the GP prior and the likelihood allows the posterior distribution
and marginal likelihood to be computed analytically. However, for
classification and other tasks involving non-Gaussian likelihoods,
this conjugacy is lost, and exact Bayesian inference generally
becomes intractable. Consequently, approximate inference methods
are required to estimate the posterior distribution and learn
the model hyperparameters.

Variational GP methods \citep{Opper2009} address this intractability by introducing a variational distribution $q_\theta(f)\approx p(f\,|\,\cal D)$ parameterised by $\theta$ and optimised by maximising the evidence lower bound (ELBO) on the log marginal likelihood $\log p(\bm y)$,
\begin{equation}
\mathrm{ELBO}(\theta)
\coloneqq
\bb E_{q_\theta(f)}
\left[
\log p(\bm y\,|\,f)
\right]
-
\mathrm{KL}\left(q_\theta(f)\,\big\|\,p(f)\right),
\end{equation}
where $\bm y\coloneqq(y_1,\ldots,y_m)$, $p(\bm y\,|\,f)$ is the product of the likelihoods $p(y_i\,|\,f(\cal K_i))$, and $\mathrm{KL}(\cdot\,\|\,\cdot)$ denotes the Kullback--Leibler divergence. The variational parameters $\theta$, the kernel and likelihood hyperparameters, and the wavelet parameters of the Hodgelet representations can be optimised jointly by maximising the ELBO; in our experiments, the KL term is down-weighted by a factor $\gamma$ (see Appendix~\ref{appendix: experiments}). This provides a principled objective for learning continuous model parameters without requiring a separate validation set for their optimisation. 

In all our experiments, we use sparse variational GPs, which introduce a set of $M$ inducing variables and reduce the cost of inference from $\cal O(m^3)$ to $\cal O(m_bM^2+M^3)$ per mini-batch of size $m_b$, which is linear in $m$ per pass over the data \citep{Titsias2009,hensman2013gaussian}; the values of $M$ are given in Appendix~\ref{appendix: experiments}.

\section{Experimental details} \label{appendix: experiments}
In this appendix, we provide additional details regarding the experiments described in Section~\ref{sec:exps}. 

For each dataset, we perform five independent repetitions. A repetition consists of a freshly generated dataset for Div-curl-free and Vortices, a fresh random subsample for ieee24, a different random split into folds for TUDataset and MoleculeNet, or a fresh balanced subsample with fresh random triangle features for MANTRA. The reported means and standard errors are computed over these five repetitions.

We implement our models in PyTorch \citep{pytorch2019}, using GPyTorch \citep{gpytorch2018} to train the Gaussian processes and PyTorch Geometric \citep{pygeometric2019} to process graph data; the Div-curl-free datasets are generated with JAX.

\subsection{Preprocessing}  \label{appendix: preprocessing}
The TUDataset graphs are used as provided. For MoleculeNet, molecules that cannot be parsed or that have fewer than three atoms are removed; the vertex attributes one-hot encode the nine atom features and the edge attributes the three bond features (bond type, stereochemistry, conjugation) computed from the SMILES strings by PyTorch Geometric, whereas for TUDataset they encode the atom and bond types. For ieee24, the transmission lines and transformers that are out of service in a sample, whose features are all zero, are removed from its graph, and parallel lines between the same pair of buses are merged by averaging their features. MANTRA has no native attributes, so we use the vertex degrees, the number of triangles incident to each edge, and eight random features on each triangle, drawn independently from the standard normal distribution and redrawn in each repetition (the GPs, SNN, and SCNN use the triangle features, and the Transformer uses the vertex and edge signals). Finally, as mentioned in Section~\ref{sec:powergraph}, the ieee24 datasets have been downsampled from $21500$ graphs, with a new subsample in each repetition. Table~\ref{tab:datasets} summarises the sizes of all datasets except MANTRA.

\begin{table}[ht]
\centering
\caption{Dataset statistics: number of complexes per repetition and mean numbers of vertices, edges, and triangles per complex, after lifting every $3$-clique to a $2$-simplex on the graph benchmarks. For Div-curl-free and Vortices, Figure~\ref{fig:divcurlfreevortices} uses $100$ complexes per generated dataset and Appendix~\ref{appendix:additional} uses $200$ unless stated otherwise; their mesh statistics are for the meshes with $n=100$ used in Appendix~\ref{appendix:additional}, which have $132$ vertices once the points on the sides are added (Appendix~\ref{appendix:vectorfield}); Figure~\ref{fig:divcurlfreevortices} varies the number of vertices from $10$ to $150$.}
\label{tab:datasets}
\begin{tabular}{llcccc}
\toprule
Dataset & Task & Complexes & Vertices & Edges & Triangles \\
\midrule
Div-curl-free & Binary & 100 / 200 & 132.0 & 357.0 & 226.0 \\
Vortices & Binary & 100 / 200 & 132.0 & 357.0 & 226.0 \\
MUTAG & Binary & 188 & 17.9 & 19.8 & 0.0 \\
AIDS & Binary & 2000 & 15.7 & 16.2 & 0.0 \\
FreeSolv & Regression & 628 & 8.9 & 8.6 & 0.0 \\
ESOL & Regression & 1121 & 13.4 & 13.8 & 0.0 \\
ieee24-binary & Binary & $\approx 376$ & 24.0 & 33.1 & 0.9 \\
ieee24-multiclass & Multiclass (4 classes) & $\approx 351$ & 24.0 & 33.0 & 0.9 \\
ieee24-regression & Regression & $\approx 123$ & 24.0 & 33.0 & 1.0 \\
\bottomrule
\end{tabular}

\end{table}

\subsection{Vector field classification experiments} 

\label{appendix:vectorfield}

Here, we give more details about the vector field classification experiments (Div-curl-free and Vortices) from Section~\ref{exp:vectorfield}.

\subsubsection{Discretising vector fields via the de Rham map}

\begin{figure}[h]
    \centering
    \subfigure[Vector field.]{{\setlength{\fboxsep}{0pt}\setlength{\fboxrule}{0.6pt}\fbox{\includegraphics[width=6.7cm, height=6.7cm, trim=56.5 31.8 63.2 45.5, clip]{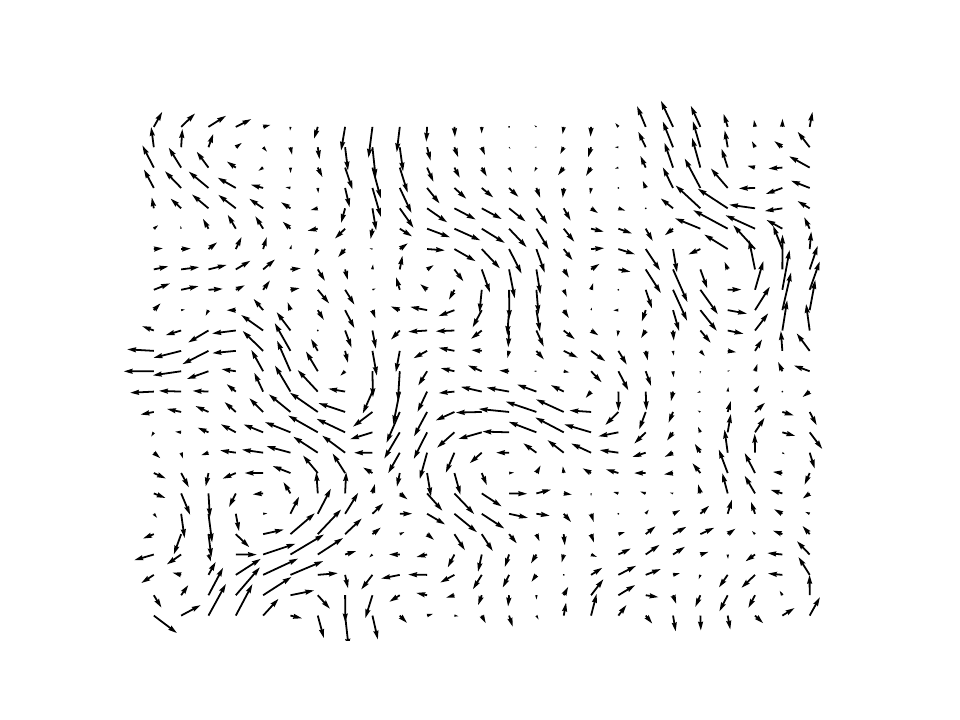}}}
    }
    \hfill
    \subfigure[Projection onto a triangular mesh.]{{\setlength{\fboxsep}{0pt}\setlength{\fboxrule}{0.6pt}\fbox{\includegraphics[width=6.7cm, height=6.7cm]{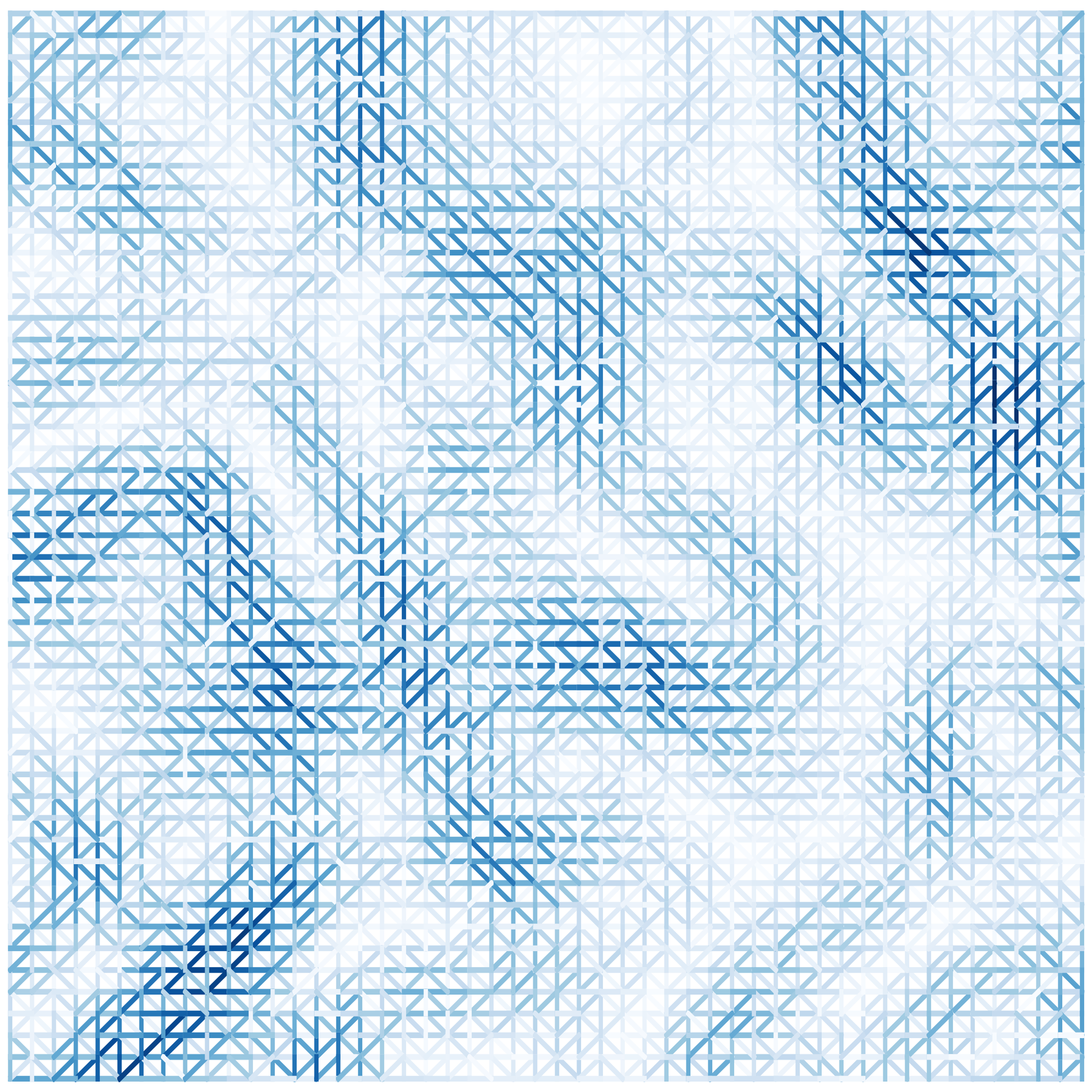}}}
    }
    \caption{Projection of a continuous vector field onto a regular triangular mesh by applying the de Rham map.}
    \label{fig:div-free-vector-field}
\end{figure}

To project a vector field $\bm X : \R^2 \rightarrow \R^2$ onto a simplicial $2$-complex, we employ the \emph{de Rham map} \citep{Desbrun2008}, which ``discretises'' a vector field into edge signals. For a given oriented edge $e = (v_0, v_1)$ with endpoint coordinates $\bm p_0$ and $\bm p_1$, we define the projection $(\bm x_1)_e$ of $\bm X$ onto $e$ by the following integral,
\begin{align}
    (\bm x_1)_e \coloneqq \int^1_0 \bm X\big(\bm p_0 + t (\bm p_1 - \bm p_0)\big) \cdot (\bm p_1 - \bm p_0) \, \mathrm{d}t,
\end{align}
that is, the line integral of $\bm X$ along the edge. This one-dimensional integral is computed by numerical quadrature. Note that this depends on the ordering of the vertices $v_0$ and $v_1$ characterising the edge: if we flip the order, then the sign of $(\bm x_1)_e$ flips. Consequently, the mesh must be oriented for the edge signal $\bm x_1$ to be well defined. In Figure~\ref{fig:div-free-vector-field}, we display an example of such a projection onto a regular triangular mesh. The meshes of our datasets are Delaunay triangulations of the square $[-1,1]^2$: a mesh of $n$ vertices, as we call it throughout, is built from the four corners and $n-4$ uniformly random interior points, with $4(\lfloor\sqrt n\rfloor-2)$ evenly spaced points added on its sides, which gives exactly $n+4(\lfloor\sqrt n\rfloor-2)$ vertices ($132$ for $n=100$).

\subsubsection{Div-curl-free}

To generate a random vector field, we first sample a Gaussian process $f : \R^2 \rightarrow \R$ and then take its derivatives $f_x, f_y$. Since a curl-free 2D vector field is the gradient of a potential, we obtain a curl-free field by taking
\begin{align}
    \bm X_{\text{curl-free}} \coloneqq \grad f = [f_x, f_y]^\top.
\end{align}

Similarly, since a divergence-free 2D vector field is the rotated gradient of a streamfunction, we obtain a divergence-free field by taking
\begin{align}
    \bm X_{\text{div-free}} \coloneqq \grad^\perp f = [f_y, -f_x]^\top.
\end{align}

\begin{figure}[htbp]
    \centering
    \subfigure[Horizontal derivative $f_x$ of a GP sample.]{\includegraphics[width=0.475\textwidth]{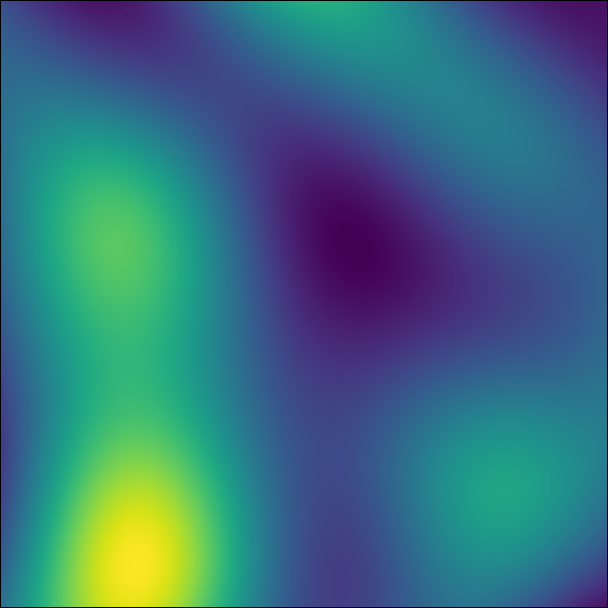}
        \label{fig:mean and std of net14}
    }\hfill
    \subfigure[Vertical derivative $f_y$ of a GP sample.]{\includegraphics[width=0.475\textwidth]{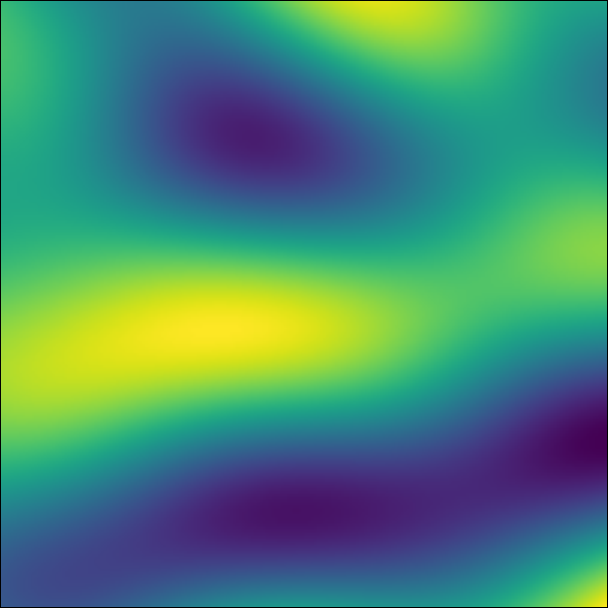}
        \label{fig:mean and std of net24}
    }\\[1ex]
    \subfigure[Divergence-free field $(f_y, -f_x)$ over $f$.]{\includegraphics[width=0.475\textwidth]{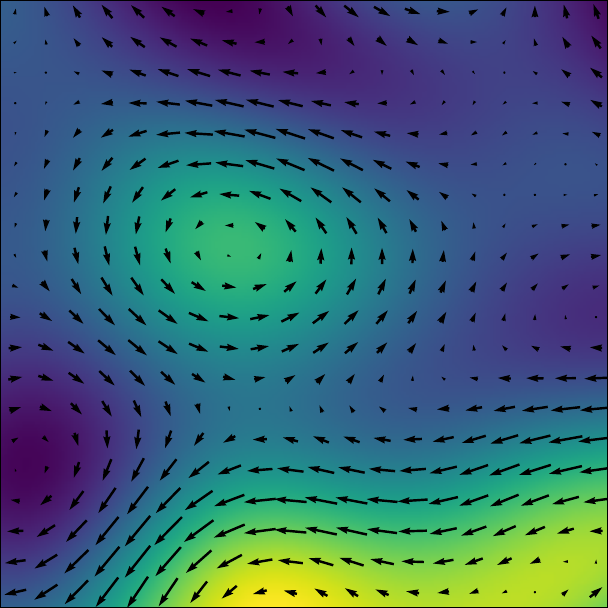}
        \label{fig:div_free}
    }\hfill
    \subfigure[Curl-free field $(f_x, f_y)$ over $f$.]{\includegraphics[width=0.475\textwidth]{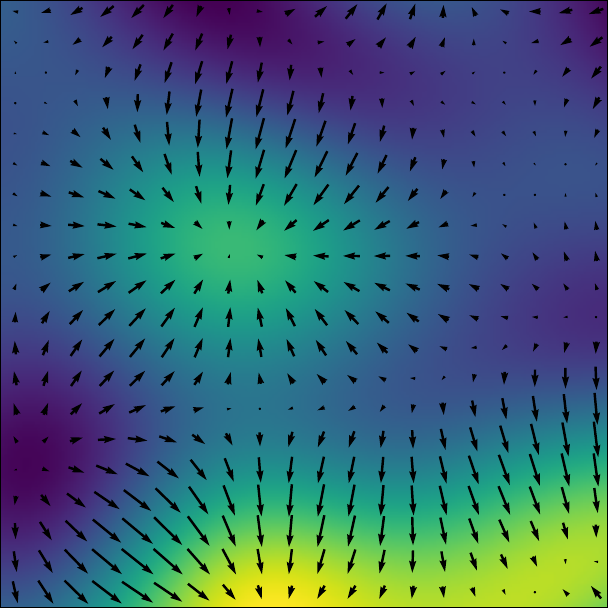}
        \label{fig:curl_free}
    }
    \caption{Illustration of the process that generates the random vector fields of Div-curl-free.}
    \label{fig:vector-field-div-free-curl-free}
\end{figure}

For the scalar field $f$, we use samples from a squared-exponential Gaussian process with lengthscale $0.5$, drawn using its random Fourier feature approximation with $1000$ cosine and $1000$ sine features \citep{rahimi2007random}. Figure~\ref{fig:vector-field-div-free-curl-free} shows the derivatives $f_x$ and $f_y$ of such a sample and the resulting divergence-free and curl-free fields.

    Our data consist of
    mostly divergence-free or mostly curl-free vector fields, which are generated by considering the linear combination
    \begin{align}
        \bm X \coloneqq \rho \bm X_{\text{div-free}} + (1 - \rho) \bm X_{\text{curl-free}},
    \end{align}
    where $\rho$ is drawn uniformly from $[0.15, 0.40]$ or from $[0.60, 0.85]$, with equal probability, and $\bm X_{\text{div-free}}$ and $\bm X_{\text{curl-free}}$ are obtained from the same GP sample $f$ as above. The edge signal of a complex is the de Rham map of $\bm X$ on its mesh, with independent Gaussian noise of standard deviation $0.5$ added to the value on each edge. The randomness of the dataset then comes from $f$, $\rho$, the noise, and the mesh, which is new for every complex. The label indicates whether $\bm X$ is mostly curl-free ($\rho<0.5$) or mostly divergence-free ($\rho>0.5$). 

\subsubsection{Vortices}
We generate vortex-dominated flows with the point-vortex model of fluid flow \citep{newton2002n}: $N_{\mathrm{v}}$ point vortices with strengths $\Gamma_1, \ldots, \Gamma_{N_{\mathrm{v}}}$ at locations $\bm p_1, \ldots, \bm p_{N_{\mathrm{v}}}$. To avoid singular velocities at the vortex locations, we use vortex-blob regularisation \citep{beale1985high} with smoothing parameter $\delta=0.1$, which gives the \emph{streamfunction}
\begin{align}
    \psi_\delta(\bm p) \coloneqq -\sum_{n=1}^{N_{\mathrm{v}}} \frac{\Gamma_n}{4\pi} \left[\log \left(\|\bm p - \bm p_n\|^2\right) - \mathrm{Ei}(-\|\bm p - \bm p_n\|^2/\delta^2)\right],
\end{align}
where $\mathrm{Ei}$ denotes the exponential integral. The velocity field at a point $\bm p=(x,y)$ is
\begin{align}
    \bm X(\bm p) \coloneqq \Big(\frac{\partial \psi_\delta}{\partial y}(\bm p), -\frac{\partial \psi_\delta}{\partial x}(\bm p)\Big)^\top,
\end{align}
as for the divergence-free fields of Div-curl-free. Its net circulation, that is, its circulation around a closed curve that encloses all vortices far from them, is
\begin{align}\label{eq:net-circulation}
    \Gamma \approx \sum_{n=1}^{N_{\mathrm{v}}} \Gamma_n,
\end{align}
with equality when $\delta=0$.

\begin{figure}[h]
    \centering
    \subfigure[Fluid flow with $3$ vortices.]{\includegraphics[height=6.7cm]{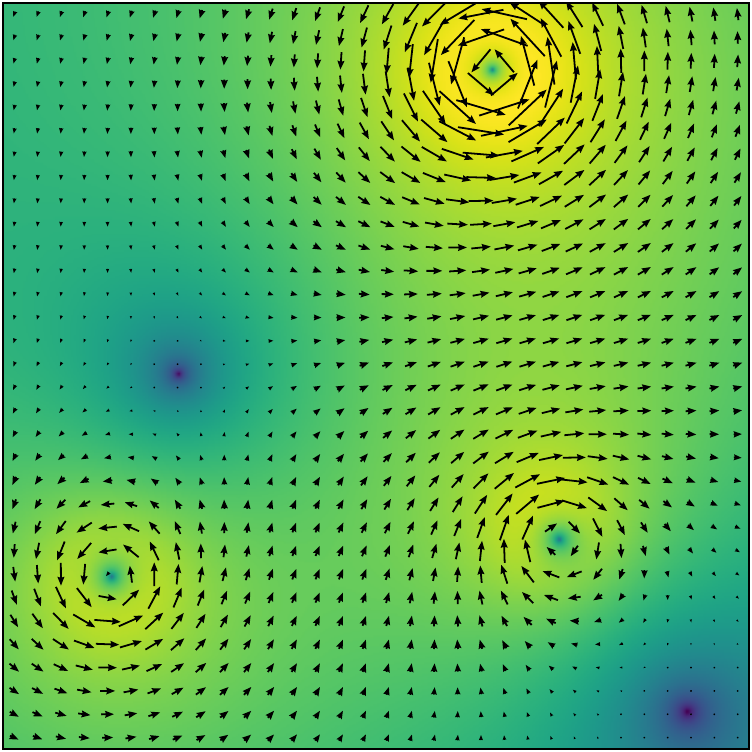}
    }
    \hfill
    \subfigure[Projection onto a triangular mesh.]{\includegraphics[height=6.7cm]{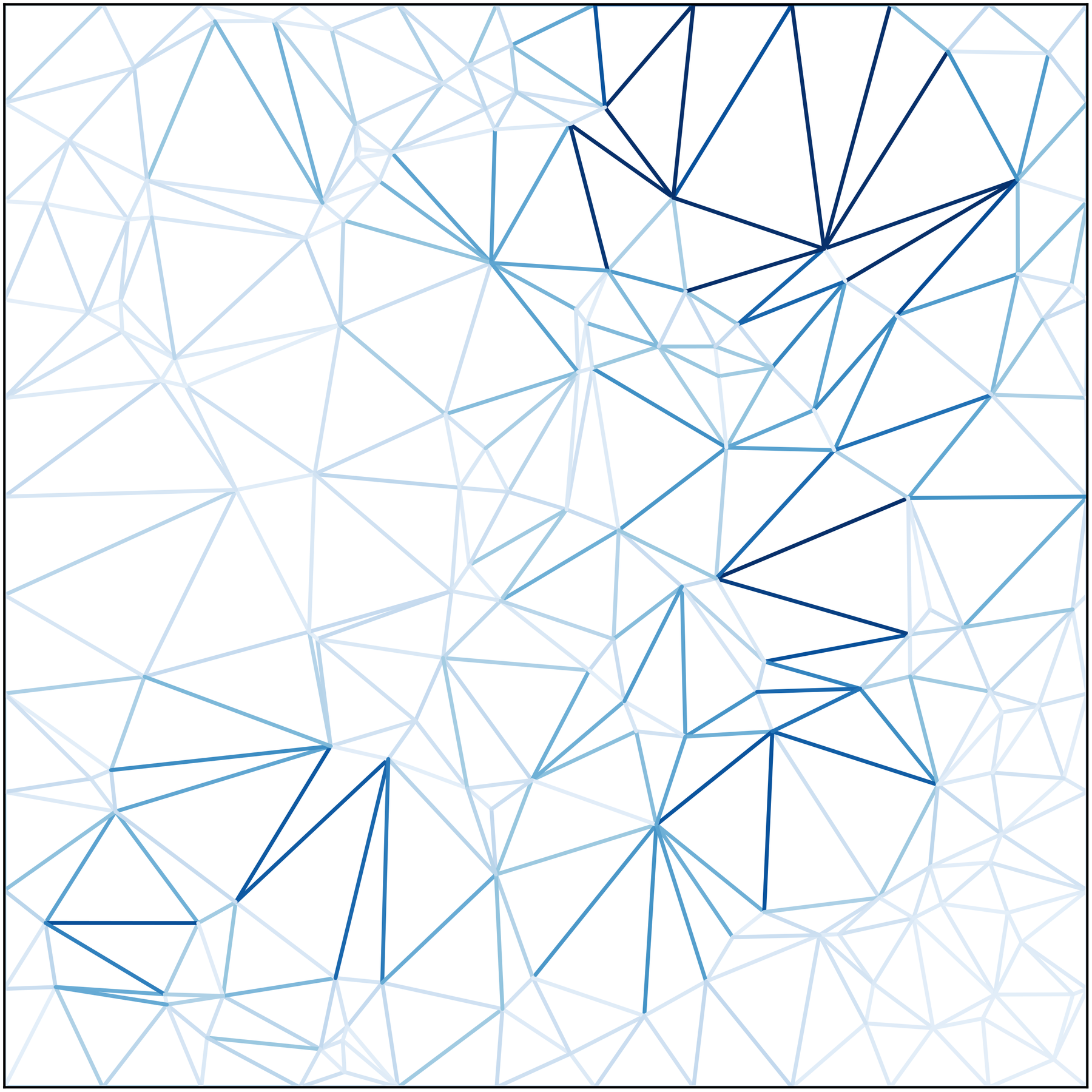}
    }
    \caption{A data point of Vortices: (a) the flow, with the colour giving its speed (log scale), and (b) its projection onto the mesh of its dataset ($150$ vertices), with the shade of an edge giving the absolute value of the edge signal.}
    \label{fig:vortices}
\end{figure}

In our experiments, $N_{\mathrm{v}}=3$, and every complex draws new vortex locations, independently and uniformly in the meshed square $[-1,1]^2$, and new strengths, independently from the standard normal distribution. Each generated dataset consists of $100$ attributed SCs ($200$ for the experiments in Appendix~\ref{appendix:additional}), obtained by applying the de Rham map to the resulting vector fields on one randomly generated triangular mesh shared by all its complexes (Figure~\ref{fig:vortices}); each repetition draws a new mesh. The label is ``0'' if $\sum_{n=1}^{N_{\mathrm{v}}}\Gamma_n \geq 0$ and ``1'' otherwise.

\paragraph{Edge orientation.} On both datasets, an edge is oriented from its lower-indexed to its higher-indexed vertex, with the vertices numbered in the order in which they are generated. This order is arbitrary, so the orientation differs from mesh to mesh. Since the edge signal of Vortices changes sign when the orientation is reversed, its sum is comparable only across complexes that share an orientation, which holds here because the complexes of a generated dataset share one mesh. Div-curl-free has a new mesh for every complex, but its label does not depend on the orientation, and the energy is invariant under it.

\subsection{Training and evaluation}
\label{appendix:training_model_selection_eval}

\paragraph{Overall pipeline.} For each SC, we (i) compute the eigendecompositions of its Hodge Laplacians once, split according to the Hodge subspaces; (ii) apply the learnable wavelet filters to the attributes and aggregate the Hodgelet coefficients into the representations $\bm r_{k\bullet}$; and (iii) evaluate the kernel \eqref{eq:add-kernel} on these representations. The wavelet filter scales and amplitudes, mixture weights (when the mixture aggregation is used), kernel lengthscales and output scales, constant mean, inducing-point locations, variational parameters, and likelihood parameters are optimised jointly by maximising the $\gamma$-weighted ELBO with stochastic gradients; only the eigendecompositions are precomputed, and gradients flow from the objective back into the representations. The remaining hyperparameters are fixed for each benchmark (see below) rather than selected within the evaluation protocol. Predictions for a new SC follow from the posterior predictive distribution \eqref{eq:gp-posterior}, with the posterior replaced by its variational approximation.

Within each of the five repetitions described at the beginning of this appendix, we run five-fold cross-validation, stratified for classification tasks. For each fold, a model is trained on the four remaining folds and predicts the held-out fold, so that every complex receives exactly one out-of-fold prediction. The metrics (accuracy or mean squared error, NLPD, and ECE) are computed once from these pooled out-of-fold predictions. We report the mean and the standard error over the five repetitions, that is, the sample standard deviation divided by $\sqrt5$; the standard error is never computed across folds, which are correlated within a repetition. For TUDataset and MoleculeNet, the repetitions reuse the same graphs with different splits, so their standard errors reflect the variability of the splits and of the training, but not that of the data. For classification, the ECE \citep{guo2017calibration} uses ten equal-width bins of the predicted probability of the positive class, except on ieee24 in Table~\ref{tab:calibration_ieee24}, where it uses the probability of the predicted class.

The GPs are trained on the full training folds for a fixed number of epochs, without early stopping. The neural baselines hold out $1/9$ of the training folds as a validation split for early stopping (patience of $50$ epochs).

\subsection{Model configurations}

In Tables~\ref{gnn_configs1} and~\ref{gnn_configs2}, we record the hyperparameters used for the GNN baselines on each dataset. All GNNs use the Adam optimiser and a batch size of $32$, map the vertex attributes to the hidden dimension with a linear layer, apply the message-passing layers with ReLU activations and dropout, pool the vertex embeddings with a global sum or mean readout, and predict with a two-layer MLP. We use the PyTorch Geometric layers: GIN uses the edge attributes as in \citet{Hu2020} (GINE), GAT and Transformer use them in their attention mechanism, with four heads, and GCN uses the vertex attributes only; all other layer settings are the PyTorch Geometric defaults. Regression targets are standardised with the statistics of the training folds, excluding the validation split. The baselines of the scaling, Div-curl-free calibration, MANTRA, and further-baseline experiments of Appendix~\ref{appendix:additional} are configured as in Table~\ref{tab:baseline_configs}.

All our Gaussian process models use the additive kernel \eqref{eq:add-kernel} (for WTGP, with one term per simplex dimension), whose component kernels $\kappa_{k\bullet}$ are radial basis function (RBF) kernels with one lengthscale per feature. The GPs have a constant mean, and regression targets are standardised with the statistics of the training folds. Likewise, each entry of the Hodgelet representations is standardised with its mean and standard deviation over the training fold, computed once at initialisation; this keeps the RBF kernels well conditioned, since the energy and the sum of the Hodgelet coefficients add one term per simplex and therefore grow with the size of the complex, but does not make the representations invariant to size. The inducing points are initialised by $k$-means on the standardised training representations. The likelihood is Bernoulli with a probit link for binary tasks, a softmax over one latent GP per class for multiclass tasks, and Gaussian for regression.

For the Hodgelet representations, we use, for all $k$, the scaling function $a_k(x)=e^{-x}$, the wavelet function $b_k(x)=x\,e^{1-x}$, which peaks at $b_k(1)=1$, and $F_k=3$ wavelet filters with $J_k=3$ band-pass scales each; the scales $\alpha_k$ and $\beta_{kl}$ of the wavelet filters (Section~\ref{sec:RH_transform}) are initialised uniformly in $[4, 6]$ and $[0.1, 5]$, respectively, and the filter amplitudes $\eta_k$ and $\eta_{kl}$ at $1$.

We use sparse variational GPs with $M=400$ inducing points on FreeSolv and ESOL; on MUTAG and ieee24, $M$ is the size of a training fold, up to $300$: $150$ or $151$ on MUTAG, $296$ to $300$ on ieee24-binary, $276$ to $286$ on ieee24-multiclass, and $96$ to $100$ on ieee24-regression. On MUTAG and ieee24-binary, the accuracy still improves up to these values (see Table~\ref{tab:inducing}). We use $M=40$ on the vector-field datasets and MANTRA, and on AIDS, where all models are close to perfect accuracy. Unless stated otherwise, all GPs are trained for $200$ epochs ($400$ on FreeSolv and $700$ on ieee24-regression) with batch size $32$.

\paragraph{Lengthscale initialisation.} The lengthscales start at the GPyTorch default of $\ln 2$, except on FreeSolv and ESOL. On these two molecular regression benchmarks, this default leaves the vertex kernels almost diagonal at initialisation. We therefore initialise all lengthscales of each component kernel to the median pairwise Euclidean distance between the standardised representations of the training complexes for that component, computed once per training fold at the initial wavelet parameters \citep[the median heuristic;][]{Garreau2017}. The lengthscales are then optimised as usual.

\paragraph{Aggregation functions.} The aggregation function is fixed per benchmark. We consider the energy, the sum, the mixture, and, for attributes that are not cochains, the orientation-averaged sum and energy \eqref{eq:oavg}. For the Hodgelet coefficients $\hat{\bm x}_{k\bullet}\in\R^{N_k}$, $\bullet\in\{e,c,h\}$, the energy is $\|\hat{\bm x}_{k\bullet}\|^2=\sum_{i=1}^{N_k}(\hat{\bm x}_{k\bullet})_i^2$, the sum is $\sum_{i=1}^{N_k}(\hat{\bm x}_{k\bullet})_i$, and the mixture is the learnable weighted combination
\begin{equation*}
    \omega_1\sum_{i=1}^{N_k}(\hat{\bm x}_{k\bullet})_i+\omega_2\|\hat{\bm x}_{k\bullet}\|^2+\omega_3\max_{i}\,(\hat{\bm x}_{k\bullet})_i+\omega_4\min_{i}\,(\hat{\bm x}_{k\bullet})_i,
\end{equation*}
with mixture weights $\omega_1,\ldots,\omega_4\in\R$, initialised at $1/4$. The wavelet coefficients of WTGP are aggregated in the same way. On the vector fields, we use the energy for Div-curl-free and the sum for Vortices, whose label is the sign of the net circulation; the energy does not contain this sign and performs poorly on Vortices (see Table~\ref{tab:aggregation}). For TUDataset and MoleculeNet, we use the mixture for vertex attributes and the orientation-averaged sum and energy \eqref{eq:oavg} for the bond attributes, which are not cochains. For ieee24, all graphs share the grid and hence one orientation, which allows us to use the mixture throughout. For MANTRA, we use the mixture aggregation on the triangle features. The ablations in Appendix~\ref{appendix:additional} use the aggregation of the corresponding benchmark.

\paragraph{Optimisation.} To optimise the ELBO, we introduce a scaling constant $\gamma$ that balances the expected log-likelihood term against the Kullback--Leibler (KL) regularisation term. Dividing the objective by $\gamma$ shows that this corresponds to variational inference with the tempered likelihood $p(\bm y\,|\,f)^{1/\gamma}$; when $\gamma<1$, this reduces the regularising effect of the prior. We use $\gamma=0.1$ for all our GP models. For training, we use the AdamW optimiser (AMSGrad variant), with learning rate $0.01$ for the wavelet, mixture, kernel, mean, and inducing-point parameters, $0.001$ for the variational and likelihood parameters, weight decay $10^{-3}$, and gradient clipping at norm $1$.

\begin{table}[ht] 
\centering
\caption{Configurations used for the graph neural network baselines in Table~\ref{TUDatasets_MoleculeNet}.}
\label{gnn_configs1}
\begin{tabular}{lccccccc}
\toprule
 & MUTAG & AIDS & FreeSolv & ESOL \\
\midrule
Hidden dim. & $128$  & $128$ & $64$ & $32$  \\ 
Num.\ layers & $5$ & $5$ & $3$ & $5$  \\
Dropout & $0.5$ & $0.5$ & $0$ & $0$ \\
Weight decay & $5\cdot10^{-5}$ & $5\cdot10^{-5}$ & $5\cdot10^{-5}$ & $5\cdot10^{-5}$ \\
Readout & Sum & Sum & Mean & Sum  \\
Learning rate & $0.01$ & $0.01$ & $0.01$ & $0.01$  \\
Max.\ epochs & $200$ & $200$ & $300$ & $200$\\
Task & Binary & Binary & Regression & Regression \\
\bottomrule
\end{tabular}
\end{table}

\begin{table}[ht] 
\centering
\caption{Configurations used for the graph neural network baselines in Table~\ref{ieee24}.}
\label{gnn_configs2}
\begin{tabular}{lccccccc}
\toprule
 & ieee24-binary & ieee24-multiclass & ieee24-regression \\
\midrule
Hidden dim. & $32$ & $32$ & $32$ \\ 
Num.\ layers & $3$ & $3$ & $3$ \\
Dropout & $0$ & $0$ & $0$ \\
Weight decay & $5\cdot10^{-5}$ & $5\cdot10^{-5}$ & $5\cdot10^{-5}$ \\
Readout & Sum & Sum & Mean \\
Learning rate & $0.001$ & $0.001$ & $0.001$ \\
Max.\ epochs & $200$ & $200$ & $200$ \\
Task & Binary & Multiclass & Regression \\
\bottomrule
\end{tabular}
\end{table}

\begin{table}[ht]
\centering
\caption{Configurations of the Transformer in the scaling, Div-curl-free calibration, and MANTRA experiments, and of the further baselines of Tables~\ref{tab:baselines} and~\ref{tab:mantra}. The neural baselines hold out $1/9$ of the training fold for early stopping (patience of $50$ epochs) and use single precision.}
\label{tab:baseline_configs}
\small\linespread{1.15}\selectfont
\begin{tabular}{p{1.9cm}p{6.3cm}p{4.3cm}}
\toprule
Method & Model & Training \\
\midrule
Transformer & Architecture described above, hidden dimension $32$, three layers, no dropout, mean readout & Adam, learning rate $10^{-3}$, weight decay $5\cdot10^{-5}$, batch size $32$, at most $200$ epochs \\
\addlinespace
SNN & Two layers of width $32$ with ReLU, polynomial filters of order $3$ in $\bm L_1$ ($\bm L_2$ on MANTRA), scaled by its largest eigenvalue, sum readout, linear head & Adam, learning rate $10^{-2}$, weight decay $10^{-4}$, full batch, at most $200$ epochs \\
\addlinespace
SCNN & As SNN, with an identity term and separate filters of order $2$ in $\bm L_1^{\mathrm{low}}$ and $\bm L_1^{\mathrm{up}}$ (on MANTRA, in $\bm L_2^{\mathrm{low}}=\bm L_2$ only, since $\bm L_2^{\mathrm{up}}=\bm 0$), each scaled by its largest eigenvalue & As SNN \\
\addlinespace
WL-SVM & Weisfeiler--Lehman subtree features ($3$ iterations, label counts normalised to unit length), linear kernel, SVM with $C=1$ and Platt-scaled probabilities & -- \\
\addlinespace
WL-GP & Features of WL-SVM, scaled linear kernel, constant mean, Bernoulli likelihood, $M=40$ inducing points initialised by $k$-means, KL term not down-weighted ($\gamma=1$) & Adam, learning rate $10^{-2}$, $300$ full-batch epochs, no early stopping \\
\bottomrule
\end{tabular}
\end{table}

\subsection{Compute details}

All reported results were computed on a single Apple M3 machine (8 CPU cores, 16\,GB of memory), on CPU, in double precision, except for the graph and simplicial neural baselines, which use single precision. A full five-repetition evaluation of one GP configuration takes a few minutes (MUTAG, and the vector-field datasets of Figure~\ref{fig:divcurlfreevortices} at $100$ vertices), up to about twenty minutes (ieee24-regression, ieee24-binary), up to an hour (ieee24-multiclass), and up to about two and a half hours (AIDS, FreeSolv, and ESOL), with several evaluations running in parallel; see Table~\ref{tab:timing} for timings.

\section{Additional experiments}\label{appendix:additional}

All accuracy and calibration results in this appendix follow the evaluation protocol of Appendix~\ref{appendix:training_model_selection_eval}: five independent repetitions, five-fold cross-validation with pooled out-of-fold predictions, and the mean and standard error over the repetitions. Unless stated otherwise, the vector-field experiments use $200$ complexes on meshes with $100$ vertices, and ieee24-binary uses about $376$ graphs, with the same configuration as in Table~\ref{ieee24}.

\subsection{Calibration of the main benchmarks}
Tables~\ref{tab:calibration_tu}, \ref{tab:calibration_ieee24}, and \ref{tab:nlpd_regression} report the NLPD and ECE of the models in Tables~\ref{TUDatasets_MoleculeNet} and~\ref{ieee24}, computed from the same pooled out-of-fold predictions. For classification, the NLPD and ECE of the neural baselines are computed from their softmax probabilities. For regression, they only output point predictions, so we use a Gaussian predictive distribution centred on the point prediction, with variance equal to the mean squared error on the validation split used for early stopping in that fold. For all models, the ECE for regression, a measure of quantile calibration \citep{kuleshov2018accurate}, is the mean, over $p \in \{0.05, 0.15, \dots, 0.95\}$, of the absolute difference between $p$ and the fraction of targets below their predictive $p$-quantile.

On the classification benchmarks, HTGP (hybrid) or HTGP (edge) attains the lowest NLPD and ECE on MUTAG and on both ieee24 tasks; on ieee24-multiclass, WTGP (edge) has the same NLPD as HTGP (edge). For example, on MUTAG, the NLPD of HTGP (hybrid) is $0.296$, compared with $0.452$ to $0.503$ for the GNNs. On AIDS, all models have an NLPD below $0.033$ and an ECE below $0.007$. On the regression benchmarks, the NLPD of the GP models largely follows the ranking of their mean squared errors, and the GP models that use the vertex attributes attain a lower NLPD than all neural baselines. The ECE does not follow this ranking: the edge-only GPs attain the lowest ECE on FreeSolv, and HTGP (edge) on ESOL, despite their large errors, and the neural baselines, whose variance is fitted on the validation split, attain a lower ECE than the GP models on ieee24-regression ($0.041$ to $0.060$ versus $0.071$ to $0.093$).

\begin{table}[ht]
\centering
\caption{NLPD ($\downarrow$) and ECE ($\downarrow$) of the models in Table~\ref{TUDatasets_MoleculeNet} on MUTAG and AIDS. Colours indicate the {\color{BrickRed} best}, {\color{Purple} second-best}, and {\color{NavyBlue} third-best} result for each column.}
\label{tab:calibration_tu}
\setlength{\tabcolsep}{4pt}
\begin{tabular}{lcccc}
\toprule
& \multicolumn{2}{c}{MUTAG} & \multicolumn{2}{c}{AIDS} \\
\cmidrule(lr){2-3}\cmidrule(lr){4-5}
 & NLPD & ECE & NLPD & ECE \\
\midrule
GCN (vertex) & $0.467 \pm 0.020$ & $0.102 \pm 0.013$ & $\color{Purple} \mathbf{0.0229} \pm 0.0009$ & $0.0049 \pm 0.0008$ \\
GAT (hybrid) & $0.503 \pm 0.021$ & $0.091 \pm 0.016$ & $\color{BrickRed} \mathbf{0.0226} \pm 0.0007$ & $0.0049 \pm 0.0008$ \\
GIN (hybrid) & $0.452 \pm 0.023$ & $0.099 \pm 0.016$ & $\color{NavyBlue} \mathbf{0.0235} \pm 0.0009$ & $0.0057 \pm 0.0010$ \\
Transformer (hybrid) & $0.490 \pm 0.037$ & $0.086 \pm 0.008$ & $\color{NavyBlue} \mathbf{0.0235} \pm 0.0007$ & $0.0048 \pm 0.0006$ \\
\midrule
WTGP (hybrid) & $0.338 \pm 0.015$ & $0.067 \pm 0.006$ & $0.0240 \pm 0.0011$ & $\color{Purple} \mathbf{0.0031} \pm 0.0005$ \\
WTGP (vertex) & $0.424 \pm 0.008$ & $\color{NavyBlue} \mathbf{0.066} \pm 0.011$ & $0.0318 \pm 0.0009$ & $0.0051 \pm 0.0004$ \\
WTGP (edge) & $\color{NavyBlue} \mathbf{0.315} \pm 0.014$ & $0.077 \pm 0.007$ & $0.0281 \pm 0.0010$ & $\color{BrickRed} \mathbf{0.0030} \pm 0.0002$ \\
HTGP (hybrid) & $\color{BrickRed} \mathbf{0.296} \pm 0.011$ & $\color{Purple} \mathbf{0.064} \pm 0.004$ & $0.0264 \pm 0.0011$ & $\color{NavyBlue} \mathbf{0.0033} \pm 0.0003$ \\
HTGP (vertex) & $0.449 \pm 0.018$ & $0.067 \pm 0.010$ & $0.0288 \pm 0.0011$ & $0.0050 \pm 0.0004$ \\
HTGP (edge) & $\color{Purple} \mathbf{0.297} \pm 0.009$ & $\color{BrickRed} \mathbf{0.055} \pm 0.002$ & $0.0273 \pm 0.0011$ & $0.0034 \pm 0.0003$ \\
\bottomrule
\end{tabular}

\end{table}

\begin{table}[ht]
\centering
\caption{NLPD ($\downarrow$) and ECE ($\downarrow$) of the models in Table~\ref{ieee24} on ieee24-binary and ieee24-multiclass. Colours indicate the {\color{BrickRed} best}, {\color{Purple} second-best}, and {\color{NavyBlue} third-best} result for each column.}
\label{tab:calibration_ieee24}
\begin{tabular}{lcccc}
\toprule
& \multicolumn{2}{c}{ieee24-binary} & \multicolumn{2}{c}{ieee24-multiclass} \\
\cmidrule(lr){2-3}\cmidrule(lr){4-5}
 & NLPD & ECE & NLPD & ECE \\
\midrule
GCN (vertex) & $0.513 \pm 0.012$ & $0.038 \pm 0.004$ & $0.764 \pm 0.043$ & $0.054 \pm 0.010$ \\
GAT (hybrid) & $0.329 \pm 0.015$ & $\color{NavyBlue} \mathbf{0.031} \pm 0.004$ & $0.397 \pm 0.020$ & $0.033 \pm 0.004$ \\
GIN (hybrid) & $0.308 \pm 0.019$ & $0.033 \pm 0.005$ & $0.348 \pm 0.014$ & $\color{Purple} \mathbf{0.028} \pm 0.003$ \\
Transformer (hybrid) & $0.306 \pm 0.030$ & $0.037 \pm 0.007$ & $0.356 \pm 0.014$ & $0.036 \pm 0.002$ \\
\midrule
WTGP (hybrid) & $\color{NavyBlue} \mathbf{0.284} \pm 0.021$ & $\color{Purple} \mathbf{0.024} \pm 0.007$ & $\color{Purple} \mathbf{0.267} \pm 0.008$ & $\color{NavyBlue} \mathbf{0.029} \pm 0.003$ \\
WTGP (vertex) & $0.445 \pm 0.013$ & $0.038 \pm 0.008$ & $0.632 \pm 0.027$ & $0.050 \pm 0.004$ \\
WTGP (edge) & $0.339 \pm 0.041$ & $0.035 \pm 0.008$ & $\color{BrickRed} \mathbf{0.235} \pm 0.011$ & $0.030 \pm 0.004$ \\
HTGP (hybrid) & $\color{Purple} \mathbf{0.257} \pm 0.018$ & $\color{NavyBlue} \mathbf{0.031} \pm 0.003$ & $\color{NavyBlue} \mathbf{0.277} \pm 0.014$ & $0.032 \pm 0.002$ \\
HTGP (vertex) & $0.437 \pm 0.012$ & $0.038 \pm 0.006$ & $0.630 \pm 0.040$ & $0.039 \pm 0.007$ \\
HTGP (edge) & $\color{BrickRed} \mathbf{0.246} \pm 0.012$ & $\color{BrickRed} \mathbf{0.021} \pm 0.003$ & $\color{BrickRed} \mathbf{0.235} \pm 0.007$ & $\color{BrickRed} \mathbf{0.027} \pm 0.003$ \\
\bottomrule
\end{tabular}

\end{table}

\begin{table}[ht]
\centering
\caption{NLPD ($\downarrow$) and ECE ($\downarrow$) of the models in Tables~\ref{TUDatasets_MoleculeNet} and~\ref{ieee24} on the regression benchmarks. Colours indicate the {\color{BrickRed} best}, {\color{Purple} second-best}, and {\color{NavyBlue} third-best} result for each column.}
\label{tab:nlpd_regression}
{\footnotesize\setlength{\tabcolsep}{2pt}\resizebox{\textwidth}{!}{\begin{tabular}{lcccccc}
\toprule
& \multicolumn{2}{c}{FreeSolv} & \multicolumn{2}{c}{ESOL} & \multicolumn{2}{c}{ieee24-regression} \\
\cmidrule(lr){2-3}\cmidrule(lr){4-5}\cmidrule(lr){6-7}
 & NLPD & ECE & NLPD & ECE & NLPD & ECE \\
\midrule
GCN (vertex) & $1.920 \pm 0.066$ & $\color{Purple} \mathbf{0.030} \pm 0.004$ & $1.058 \pm 0.007$ & $\color{Purple} \mathbf{0.015} \pm 0.001$ & $-2.266 \pm 0.078$ & $\color{Purple} \mathbf{0.047} \pm 0.001$ \\
GAT (hybrid) & $2.047 \pm 0.053$ & $0.039 \pm 0.005$ & $1.057 \pm 0.015$ & $0.029 \pm 0.006$ & $-2.294 \pm 0.082$ & $\color{BrickRed} \mathbf{0.041} \pm 0.003$ \\
GIN (hybrid) & $1.898 \pm 0.046$ & $0.041 \pm 0.006$ & $1.393 \pm 0.048$ & $\color{Purple} \mathbf{0.015} \pm 0.003$ & $-2.398 \pm 0.045$ & $0.060 \pm 0.011$ \\
Transformer (hybrid) & $1.915 \pm 0.079$ & $0.042 \pm 0.004$ & $1.086 \pm 0.023$ & $\color{NavyBlue} \mathbf{0.018} \pm 0.002$ & $-2.380 \pm 0.042$ & $\color{NavyBlue} \mathbf{0.058} \pm 0.008$ \\
\midrule
WTGP (hybrid) & $\color{BrickRed} \mathbf{1.318} \pm 0.023$ & $0.045 \pm 0.002$ & $\color{BrickRed} \mathbf{0.939} \pm 0.010$ & $0.025 \pm 0.004$ & $\color{NavyBlue} \mathbf{-2.858} \pm 0.027$ & $0.079 \pm 0.003$ \\
WTGP (vertex) & $\color{Purple} \mathbf{1.339} \pm 0.024$ & $0.042 \pm 0.002$ & $\color{Purple} \mathbf{0.941} \pm 0.007$ & $0.027 \pm 0.004$ & $-2.649 \pm 0.047$ & $0.071 \pm 0.004$ \\
WTGP (edge) & $2.554 \pm 0.016$ & $\color{BrickRed} \mathbf{0.019} \pm 0.002$ & $1.641 \pm 0.007$ & $\color{Purple} \mathbf{0.015} \pm 0.002$ & $-2.533 \pm 0.102$ & $0.082 \pm 0.004$ \\
HTGP (hybrid) & $\color{NavyBlue} \mathbf{1.415} \pm 0.008$ & $\color{NavyBlue} \mathbf{0.032} \pm 0.001$ & $\color{NavyBlue} \mathbf{0.972} \pm 0.006$ & $0.023 \pm 0.002$ & $\color{BrickRed} \mathbf{-2.996} \pm 0.051$ & $0.088 \pm 0.009$ \\
HTGP (vertex) & $1.445 \pm 0.008$ & $0.038 \pm 0.004$ & $0.991 \pm 0.012$ & $0.028 \pm 0.005$ & $-2.645 \pm 0.032$ & $0.077 \pm 0.005$ \\
HTGP (edge) & $2.533 \pm 0.008$ & $\color{BrickRed} \mathbf{0.019} \pm 0.001$ & $1.641 \pm 0.006$ & $\color{BrickRed} \mathbf{0.014} \pm 0.004$ & $\color{Purple} \mathbf{-2.927} \pm 0.059$ & $0.093 \pm 0.007$ \\
\bottomrule
\end{tabular}
}}
\end{table}

\subsection{Aggregation function}
Table~\ref{tab:aggregation} compares the energy, sum, and mixture aggregations for the edge models.

The best aggregation depends on the task. On Div-curl-free, the energy performs best, closely followed by the mixture, whereas the sum performs poorly ($0.599$ for HTGP), since it depends on the signs of the edge values and hence on the edge orientation, which differs from complex to complex. On Vortices, the label is the sign of the circulation, which the energy discards: the models with the energy are at chance level, whereas the sum reaches $0.814$ for HTGP and $0.767$ for WTGP. On Vortices, the mixture lies between the energy and the sum. On ieee24-binary, the three aggregations perform similarly (between $0.900$ and $0.915$), except for WTGP (edge) with the mixture ($0.843$).

\begin{table}[ht]
\centering
\caption{Accuracy for different aggregation functions. The best result in each row is in bold.}
\label{tab:aggregation}
\begin{tabular}{llccc}
\toprule
Dataset & Model & Energy & Sum & Mixture \\
\midrule
Div-curl-free & HTGP (edge) & $\mathbf{0.898} \pm 0.005$ & $0.599 \pm 0.030$ & $0.888 \pm 0.005$ \\
Div-curl-free & WTGP (edge) & $\mathbf{0.823} \pm 0.019$ & $0.551 \pm 0.020$ & $0.816 \pm 0.021$ \\
\midrule
Vortices & HTGP (edge) & $0.534 \pm 0.026$ & $\mathbf{0.814} \pm 0.023$ & $0.731 \pm 0.056$ \\
Vortices & WTGP (edge) & $0.519 \pm 0.031$ & $\mathbf{0.767} \pm 0.036$ & $0.754 \pm 0.049$ \\
\midrule
ieee24-binary & HTGP (edge) & $\mathbf{0.915} \pm 0.008$ & $0.902 \pm 0.006$ & $0.902 \pm 0.007$ \\
ieee24-binary & WTGP (edge) & $0.900 \pm 0.009$ & $\mathbf{0.904} \pm 0.002$ & $0.843 \pm 0.025$ \\
\bottomrule
\end{tabular}

\end{table}

\subsection{Hodge components}
Table~\ref{tab:hodge} reports the accuracy of HTGP (edge) when only one Hodge component is used, compared with all components. The meshes have no holes, so their harmonic subspace is empty.

The benchmarks rely on different subspaces. On Div-curl-free, the exact and co-exact components alone reach $0.769$ and $0.860$, and combining them improves the accuracy to $0.898$; both carry information, since the label compares the divergence-free and curl-free parts of the field. On Vortices, the exact and co-exact components alone reach $0.728$ and $0.740$, and combining them improves the accuracy to $0.814$. On ieee24-binary, the exact component carries most of the information ($0.892$, compared with $0.902$ for all components), followed by the harmonic component ($0.850$); the co-exact component, which is supported on the few triangles of the power grids, is the least informative ($0.758$).

\begin{table}[ht]
\centering
\caption{Ablation of the Hodge components (HTGP (edge); energy on Div-curl-free, sum on Vortices, mixture on ieee24-binary). ``--'' indicates an empty subspace. The best result in each row is in bold.}
\label{tab:hodge}
\begin{tabular}{lcccc}
\toprule
Dataset & Harmonic & Co-exact & Exact & All \\
\midrule
Div-curl-free & -- & $0.860 \pm 0.014$ & $0.769 \pm 0.006$ & $\mathbf{0.898} \pm 0.005$ \\
Vortices & -- & $0.740 \pm 0.052$ & $0.728 \pm 0.016$ & $\mathbf{0.814} \pm 0.023$ \\
ieee24-binary & $0.850 \pm 0.010$ & $0.758 \pm 0.011$ & $0.892 \pm 0.010$ & $\mathbf{0.902} \pm 0.007$ \\
\bottomrule
\end{tabular}

\end{table}

\subsection{Comparison with further baselines}
Table~\ref{tab:baselines} compares the GP models with four further baselines: the simplicial neural network (SNN) of \citet{ebli2020simplicial} and the simplicial convolutional neural network (SCNN) of \citet{yang2022simplicial}, both with a sum readout, and the Weisfeiler--Lehman subtree kernel \citep{shervashidze2011weisfeiler} combined with an SVM (WL-SVM) or a GP classifier (WL-GP). The simplicial networks use the same edge signals and Hodge Laplacians as our models; on MUTAG, they therefore depend on the atom numbering through the bond attributes. The WL kernels need a discrete label on each vertex. On the vector-field datasets, we compute two numbers at each vertex, the divergence $(-\bm B_1\bm x_1)_v$ and the incident energy $(|\bm B_1|\bm x_1^2)_v$ of the edge signal, split the range of each into four intervals at its quartiles over all vertices of the dataset, and label the vertex by the pair of intervals into which its two numbers fall, which gives $16$ possible labels. On MUTAG, the labels are the atom types. Table~\ref{tab:baseline_configs} gives the configurations of all four baselines.

HTGP attains the highest accuracy and the lowest NLPD on all three datasets. On Div-curl-free, all four baselines are close to chance level, whereas HTGP and WTGP reach $0.898$ and $0.823$. On Vortices, the simplicial networks reach $0.632$ and $0.645$, compared with $0.814$ for HTGP, and the WL kernels are close to chance level, since their vertex labels do not encode the sign of the circulation. On MUTAG, the simplicial networks also fall behind, whereas the WL kernels perform reasonably.

\begin{table}[ht]
\centering
\caption{Comparison with further baselines. The vector-field datasets use the edge signals, with the energy on Div-curl-free and the sum on Vortices for the GP models; on MUTAG, the GP models use vertex and edge attributes. Colours indicate the {\color{BrickRed} best}, {\color{Purple} second-best}, and {\color{NavyBlue} third-best} result for each dataset and metric.}
\label{tab:baselines}
\begin{tabular}{llccc}
\toprule
Dataset & Method & NLPD ($\downarrow$) & ECE ($\downarrow$) & Accuracy ($\uparrow$) \\
\midrule
Div-curl-free & HTGP (edge) & $\color{BrickRed} \mathbf{0.245} \pm 0.008$ & $\color{Purple} \mathbf{0.041} \pm 0.006$ & $\color{BrickRed} \mathbf{0.898} \pm 0.005$ \\
Div-curl-free & WTGP (edge) & $\color{Purple} \mathbf{0.399} \pm 0.038$ & $0.053 \pm 0.005$ & $\color{Purple} \mathbf{0.823} \pm 0.019$ \\
Div-curl-free & SNN & $0.769 \pm 0.025$ & $0.145 \pm 0.016$ & $\color{NavyBlue} \mathbf{0.516} \pm 0.013$ \\
Div-curl-free & SCNN & $0.727 \pm 0.006$ & $0.117 \pm 0.009$ & $0.509 \pm 0.011$ \\
Div-curl-free & WL-SVM & $0.696 \pm 0.002$ & $\color{NavyBlue} \mathbf{0.045} \pm 0.013$ & $0.481 \pm 0.013$ \\
Div-curl-free & WL-GP & $\color{NavyBlue} \mathbf{0.693} \pm 0.001$ & $\color{BrickRed} \mathbf{0.025} \pm 0.010$ & $0.500 \pm 0.016$ \\
\midrule
Vortices & HTGP (edge) & $\color{BrickRed} \mathbf{0.434} \pm 0.043$ & $\color{NavyBlue} \mathbf{0.058} \pm 0.009$ & $\color{BrickRed} \mathbf{0.814} \pm 0.023$ \\
Vortices & WTGP (edge) & $\color{Purple} \mathbf{0.510} \pm 0.055$ & $0.071 \pm 0.011$ & $\color{Purple} \mathbf{0.767} \pm 0.036$ \\
Vortices & SNN & $0.652 \pm 0.017$ & $0.105 \pm 0.013$ & $0.632 \pm 0.018$ \\
Vortices & SCNN & $\color{NavyBlue} \mathbf{0.618} \pm 0.024$ & $0.090 \pm 0.014$ & $\color{NavyBlue} \mathbf{0.645} \pm 0.042$ \\
Vortices & WL-SVM & $0.689 \pm 0.004$ & $\color{Purple} \mathbf{0.046} \pm 0.008$ & $0.542 \pm 0.017$ \\
Vortices & WL-GP & $0.680 \pm 0.005$ & $\color{BrickRed} \mathbf{0.036} \pm 0.011$ & $0.555 \pm 0.009$ \\
\midrule
MUTAG & HTGP (hybrid) & $\color{BrickRed} \mathbf{0.296} \pm 0.011$ & $\color{BrickRed} \mathbf{0.064} \pm 0.004$ & $\color{BrickRed} \mathbf{0.878} \pm 0.009$ \\
MUTAG & WTGP (hybrid) & $\color{Purple} \mathbf{0.338} \pm 0.015$ & $\color{Purple} \mathbf{0.067} \pm 0.006$ & $\color{Purple} \mathbf{0.863} \pm 0.010$ \\
MUTAG & SNN & $0.526 \pm 0.031$ & $0.121 \pm 0.007$ & $0.732 \pm 0.017$ \\
MUTAG & SCNN & $0.532 \pm 0.049$ & $0.118 \pm 0.010$ & $0.738 \pm 0.016$ \\
MUTAG & WL-SVM & $0.429 \pm 0.002$ & $0.136 \pm 0.008$ & $0.794 \pm 0.003$ \\
MUTAG & WL-GP & $\color{NavyBlue} \mathbf{0.374} \pm 0.005$ & $\color{NavyBlue} \mathbf{0.114} \pm 0.009$ & $\color{NavyBlue} \mathbf{0.802} \pm 0.006$ \\
\bottomrule
\end{tabular}

\end{table}

\subsection{Effect of the $2$-simplices}
On the meshes, the $2$-simplices are native; on the graph benchmarks, they are added by filling every $3$-clique. The number of triangles varies widely: the meshes have $226$ triangles per complex, ieee24 about one, and MUTAG none. Table~\ref{tab:triangles} compares HTGP (edge) with and without the $2$-simplices; without them, the co-exact subspace vanishes.

Removing the $2$-simplices has little effect on ieee24-binary (from $0.902$ to $0.899$), but lowers the accuracy on Div-curl-free from $0.898$ to $0.838$ and on Vortices from $0.814$ to $0.754$. Hence, on ieee24-binary, separating the exact component from the rest of the edge signal already gives most of the benefit of the Hodge decomposition, whereas on the meshes, the co-exact subspace created by the $2$-simplices matters as well.

\begin{table}[ht]
\centering
\caption{Accuracy of HTGP (edge) with and without $2$-simplices (energy on Div-curl-free, sum on Vortices, mixture on ieee24-binary). The best result in each row is in bold.}
\label{tab:triangles}
\begin{tabular}{lcc}
\toprule
Dataset & HTGP (edge) with $2$-simplices & HTGP (edge) without $2$-simplices \\
\midrule
Div-curl-free & $\mathbf{0.898} \pm 0.005$ & $0.838 \pm 0.009$ \\
Vortices & $\mathbf{0.814} \pm 0.023$ & $0.754 \pm 0.020$ \\
ieee24-binary & $\mathbf{0.902} \pm 0.007$ & $0.899 \pm 0.009$ \\
\bottomrule
\end{tabular}

\end{table}

\subsection{Data efficiency on ieee24}
Table~\ref{tab:scaling} compares HTGP (hybrid) with the Transformer on ieee24-binary as the number of graphs $N$ grows. HTGP uses the energy aggregation, $600$ training epochs, and $M=80$ inducing points, capped at the size of a training fold ($75$ to $77$ graphs) at $N=100$. For each $N$, we draw $N/2$ graphs per class and keep only the connected graphs. HTGP is more accurate up to $N=2000$; the two models are tied at $N=2500$, and the Transformer is slightly more accurate at $N=3000$ ($0.951$ versus $0.947$). HTGP attains a lower NLPD at every size, including when the Transformer matches or exceeds its accuracy, and a lower ECE up to $N=2500$; at $N=3000$, the two ECEs are almost equal ($0.017$ versus $0.016$).

\begin{table}[ht]
\centering
\caption{Scaling with the number of graphs $N$ on ieee24-binary: HTGP (hybrid) versus the Transformer. Accuracy ($\uparrow$), NLPD ($\downarrow$), and ECE ($\downarrow$); standard errors are given for the accuracy.}
\label{tab:scaling}
\resizebox{\textwidth}{!}{\begin{tabular}{lcccccc}
\toprule
$N$ & HTGP acc. & Transf.\ acc. & HTGP NLPD & Transf.\ NLPD & HTGP ECE & Transf.\ ECE \\
\midrule
100 & $0.855 \pm 0.018$ & $0.790 \pm 0.030$ & $0.461$ & $0.499$ & $0.106$ & $0.110$ \\
500 & $0.917 \pm 0.008$ & $0.891 \pm 0.004$ & $0.270$ & $0.290$ & $0.037$ & $0.045$ \\
1000 & $0.927 \pm 0.004$ & $0.912 \pm 0.004$ & $0.205$ & $0.247$ & $0.024$ & $0.029$ \\
1500 & $0.934 \pm 0.004$ & $0.924 \pm 0.006$ & $0.184$ & $0.217$ & $0.017$ & $0.022$ \\
2000 & $0.943 \pm 0.001$ & $0.930 \pm 0.002$ & $0.159$ & $0.201$ & $0.019$ & $0.021$ \\
2500 & $0.945 \pm 0.003$ & $0.945 \pm 0.004$ & $0.149$ & $0.168$ & $0.013$ & $0.018$ \\
3000 & $0.947 \pm 0.003$ & $0.951 \pm 0.004$ & $0.143$ & $0.156$ & $0.017$ & $0.016$ \\
\bottomrule
\end{tabular}
}
\end{table}

\subsection{Calibration on Div-curl-free}
Table~\ref{tab:calibration_divcurl} reports the NLPD, ECE, and accuracy on Div-curl-free as the number of complexes $N$ grows (energy aggregation). The Transformer stays close to chance level at every dataset size (accuracy $0.494$ to $0.528$); its ECE is low only because its predicted probabilities stay close to $1/2$, as its NLPD of about $0.69$, that of always predicting $1/2$, shows. HTGP attains the lowest NLPD and the highest accuracy throughout, and a lower ECE than WTGP at every dataset size.

\begin{table}[ht]
\centering
\caption{Calibration on Div-curl-free versus the number of complexes $N$. For each $N$, the best result for each metric is in bold.}
\label{tab:calibration_divcurl}
\begin{tabular}{llccc}
\toprule
$N$ & Model & NLPD ($\downarrow$) & ECE ($\downarrow$) & Accuracy ($\uparrow$) \\
\midrule
50 & HTGP (edge) & $\mathbf{0.305} \pm 0.036$ & $0.114 \pm 0.010$ & $\mathbf{0.880} \pm 0.030$ \\
 & WTGP (edge) & $0.499 \pm 0.063$ & $0.121 \pm 0.007$ & $0.776 \pm 0.047$ \\
 & Transformer & $0.692 \pm 0.003$ & $\mathbf{0.061} \pm 0.017$ & $0.520 \pm 0.021$ \\
\midrule
100 & HTGP (edge) & $\mathbf{0.271} \pm 0.006$ & $0.057 \pm 0.005$ & $\mathbf{0.888} \pm 0.010$ \\
 & WTGP (edge) & $0.423 \pm 0.039$ & $0.092 \pm 0.012$ & $0.786 \pm 0.028$ \\
 & Transformer & $0.705 \pm 0.011$ & $\mathbf{0.027} \pm 0.009$ & $0.528 \pm 0.008$ \\
\midrule
250 & HTGP (edge) & $\mathbf{0.232} \pm 0.012$ & $0.040 \pm 0.002$ & $\mathbf{0.906} \pm 0.006$ \\
 & WTGP (edge) & $0.397 \pm 0.022$ & $0.049 \pm 0.004$ & $0.820 \pm 0.017$ \\
 & Transformer & $0.697 \pm 0.002$ & $\mathbf{0.037} \pm 0.010$ & $0.494 \pm 0.008$ \\
\midrule
500 & HTGP (edge) & $\mathbf{0.226} \pm 0.007$ & $0.029 \pm 0.001$ & $\mathbf{0.912} \pm 0.007$ \\
 & WTGP (edge) & $0.410 \pm 0.011$ & $0.039 \pm 0.003$ & $0.810 \pm 0.009$ \\
 & Transformer & $0.693 \pm 0.000$ & $\mathbf{0.011} \pm 0.001$ & $0.503 \pm 0.002$ \\
\midrule
1000 & HTGP (edge) & $\mathbf{0.213} \pm 0.004$ & $0.021 \pm 0.002$ & $\mathbf{0.915} \pm 0.004$ \\
 & WTGP (edge) & $0.400 \pm 0.003$ & $0.026 \pm 0.003$ & $0.815 \pm 0.003$ \\
 & Transformer & $0.692 \pm 0.001$ & $\mathbf{0.009} \pm 0.004$ & $0.515 \pm 0.009$ \\
\bottomrule
\end{tabular}

\end{table}

\subsection{Number of inducing points}
Table~\ref{tab:inducing} reports the accuracy of HTGP (hybrid) as a function of the number of inducing points $M$, on MUTAG and ieee24-binary with the configurations of Tables~\ref{TUDatasets_MoleculeNet} and~\ref{ieee24}. On both datasets, the accuracy increases quickly with $M$ and is within two points of its maximum from $M=50$ on. It is highest with the $M$ of Tables~\ref{TUDatasets_MoleculeNet} and~\ref{ieee24} ($150$ or $151$ on MUTAG, $296$ to $300$ on ieee24-binary): HTGP (hybrid) then reaches $0.878$ on MUTAG and $0.906$ on ieee24-binary, compared with $0.865$ at $M=100$ and $0.898$ at $M=200$, respectively.

\begin{table}[ht]
\centering
\caption{Effect of the number of inducing points $M$ on the accuracy of HTGP (hybrid). On MUTAG, $M$ is capped at the size of a training fold ($150$ or $151$ graphs); larger values are not shown.}
\label{tab:inducing}
\begin{tabular}{lcc}
\toprule
$M$ & MUTAG & ieee24-binary \\
\midrule
1 & $0.665 \pm 0.000$ & $0.605 \pm 0.020$ \\
5 & $0.810 \pm 0.003$ & $0.850 \pm 0.009$ \\
10 & $0.831 \pm 0.004$ & $0.862 \pm 0.009$ \\
50 & $0.870 \pm 0.003$ & $0.886 \pm 0.008$ \\
100 & $0.865 \pm 0.009$ & $0.894 \pm 0.007$ \\
200 & -- & $0.898 \pm 0.007$ \\
\bottomrule
\end{tabular}

\end{table}

\subsection{Training time}
Table~\ref{tab:timing} compares the training time of HTGP and the Transformer over $200$ epochs. HTGP is cheaper per epoch on the meshes of Div-curl-free, but not on the small ieee24 graphs. The eigendecomposition of the Hodge Laplacians is performed once as preprocessing and takes at most $3.6$\,s here. In terms of total time, HTGP is faster than the Transformer on Div-curl-free and about $2.5$ times slower on ieee24-binary, and the sparse variational approximation keeps its cost linear in the number of complexes.

\begin{table}[ht]
\centering
\caption{Training time of HTGP and the Transformer for $200$ epochs on one training fold (mean of three runs on one CPU thread, with no other experiments running), with the configurations of Tables~\ref{tab:scaling} and~\ref{tab:calibration_divcurl}: HTGP (hybrid) on ieee24-binary and HTGP (edge) on Div-curl-free. The total of HTGP includes the eigendecomposition for all complexes and the initialisation.}
\label{tab:timing}
\resizebox{\textwidth}{!}{\begin{tabular}{lcccc}
\toprule
Dataset & HTGP ms/epoch & Transf.\ ms/epoch & Eigendecomposition & Total (HTGP / Transf.) \\
\midrule
ieee24-binary ($N=300$, $M=80$) & $67.5$ & $27.3$ & $0.08$\,s & $13.6$\,s / $5.5$\,s \\
ieee24-binary ($N=1000$, $M=80$) & $209.8$ & $86.6$ & $0.27$\,s & $42.4$\,s / $17.3$\,s \\
Div-curl-free ($N=200$, $M=40$) & $77.3$ & $128.8$ & $3.58$\,s & $19.2$\,s / $25.8$\,s \\
\bottomrule
\end{tabular}
}
\end{table}

\subsection{Topological prediction on MANTRA}\label{appendix:mantra}
To test our models on data with native higher-order structure, we consider MANTRA \citep{Ballester2024}, a dataset of triangulations of closed surfaces, and predict whether a surface is orientable. Orientability is determined by how the triangles are glued together: for a connected closed surface, the harmonic subspace of the triangles, $\ker(\bm L_2)$, has dimension $1$ if the surface is orientable and $0$ otherwise. In each repetition, we draw a balanced sample of $500$ surfaces ($250$ per class). Since MANTRA has no attributes, we give each triangle eight random numbers, drawn independently from the standard normal distribution and redrawn in each repetition (Appendix~\ref{appendix: preprocessing}). These features carry no information about the label, but their projection onto the harmonic subspace is non-zero with probability one if the surface is orientable, and zero otherwise. HTGP (triangle) and WTGP (triangle) use only the triangle features: HTGP splits them into their exact and harmonic components (the co-exact subspace of the triangles is empty, as there are no $3$-simplices), whereas WTGP applies its wavelet filters to all eigenvalues of $\bm L_2$ at once, without separating the components. SNN and SCNN take the same triangle features as input, with polynomial filters in $\bm L_2$ instead of $\bm L_1$. The Transformer operates on the $1$-skeleton graph, with the vertex degrees and the numbers of triangles per edge as attributes. The graph carries only partial information about orientability: the Transformer reaches an accuracy of $0.744$. SNN and SCNN, whose low-order polynomial filters presumably cannot isolate the harmonic component, stay close to chance level ($0.508$ and $0.516$), whereas HTGP (triangle) and WTGP (triangle) are almost perfect ($1.000$ and $0.992$; Table~\ref{tab:mantra}).

\begin{table}[ht]
\centering
\caption{MANTRA orientability (triangulated surfaces). The best result for each metric is in bold.}
\label{tab:mantra}
\begin{tabular}{lccc}
\toprule
Method & NLPD ($\downarrow$) & ECE ($\downarrow$) & Accuracy ($\uparrow$) \\
\midrule
HTGP (triangle) & $\mathbf{0.004} \pm 0.000$ & $\mathbf{0.003} \pm 0.000$ & $\mathbf{1.000} \pm 0.000$ \\
WTGP (triangle) & $0.028 \pm 0.004$ & $0.007 \pm 0.001$ & $0.992 \pm 0.001$ \\
SNN & $0.711 \pm 0.006$ & $0.072 \pm 0.004$ & $0.508 \pm 0.011$ \\
SCNN & $0.707 \pm 0.005$ & $0.069 \pm 0.013$ & $0.516 \pm 0.003$ \\
Transformer & $0.544 \pm 0.020$ & $0.073 \pm 0.014$ & $0.744 \pm 0.018$ \\
\bottomrule
\end{tabular}

\end{table}

\end{document}